\documentclass{article}

\PassOptionsToPackage{sort&compress,numbers}{natbib}

\usepackage[final]{style/neurips_2020}

\usepackage[utf8]{inputenc} 
\usepackage[T1]{fontenc}    
\usepackage[colorlinks,allcolors=black,
            pdftex,
            pdfauthor={Steffen Schneider, Evgenia Rusak, Luisa Eck, Oliver Bringmann, Wieland Brendel and Matthias Bethge},
            pdftitle={Improving robustness against common corruptions by covariate shift adaptation},
            pdfsubject={Improving robustness against common corruptions by covariate shift adaptation},
            pdfkeywords={Covariate Shift, ImageNet-C, Domain Adaptation}
]{hyperref}
\usepackage{url}            
\usepackage{booktabs}       
\usepackage{amsfonts}       
\usepackage{nicefrac}       
\usepackage{microtype}      

\usepackage{minitoc}

\usepackage{bm}
\usepackage[dvipsnames]{xcolor}
\usepackage{svg}
\usepackage{mathtools}
\usepackage{import}
\usepackage{wrapfig}
\usepackage{multirow}
\usepackage[export]{adjustbox}

\newcommand{\Exp}{\mathbb{E}}

\newcommand{\xx}{\mathbf{x}}
\newcommand{\zz}{\mathbf{z}}

\newcommand{\normaldist}{\mathcal{N}}

\newcommand{\cov}{\bm{\Sigma}}

\newcommand{\mumu}{\bm{\mu}}

\newcommand{\covd}{\bm{\Sigma}}

\DeclareMathOperator{\Tr}{Tr}

\usepackage{tikz}

\usepackage{pgfplots}
\pgfplotsset{compat=1.13}
\pgfplotsset{
    table/search path={figs/data},
}

\usepackage{xfrac}
\usetikzlibrary{arrows}

\usetikzlibrary{math}
\usetikzlibrary{shapes,arrows}
\usetikzlibrary{positioning}
\usetikzlibrary{decorations,patterns}
\usetikzlibrary{calc,fit,matrix,chains,scopes}

\usepgfplotslibrary{colorbrewer}

\newcommand{\lw}{1pt}

\definecolor{FAIRlg}{HTML}{F7F4F0}
\definecolor{FAIRlb}{HTML}{4267B6}
\definecolor{FAIRdb}{HTML}{003462}

\definecolor{FAIRaclg}{HTML}{F4F4F4}
\definecolor{FAIRaclb}{HTML}{EDF0F6}
\definecolor{FAIRdark}{HTML}{162643}

\definecolor{accent1}{HTML}{4267B6}
\definecolor{accent2}{HTML}{003462}
\definecolor{accent3}{HTML}{162643}

\definecolor{myblue}{RGB}{31, 119, 180}
\definecolor{myred}{RGB}{44, 160, 44}
\definecolor{mygreen}{RGB}{255, 127, 14}
\definecolor{myviolet}{RGB}{140, 86, 75}

\tikzset{%
    block/.style        = {line width=\lw, rectangle, draw=black,dashed, minimum width=\dist, minimum height=.33*\dist},
    representation/.style        = {line width=\lw, rectangle, draw=white!0,fill=mygreen, minimum width=.05*\dist, minimum height=.3\dist},
    filled/.style        = {line width=\lw, rectangle, draw=white!0, rounded corners, minimum width=.15*\dist, minimum height=.05*\dist},
    line/.style         = {draw=gray, -latex, line width=.75*\lw,rounded corners},
    lineplain/.style    = {draw=gray, -, line width=.75*\lw, rounded corners},
    mathop/.style       = {draw, circle, fill=gray!20},
    branch/.style       = {fill,shape=circle,minimum size=\dist,inner sep=0pt},
    encoder/.style      = {line width=\lw, trapezium, trapezium angle=80, rotate=-90, minimum width=.55*\dist,  draw, trapezium stretches=true, fill=myblue,draw=white!0,minimum height=.15*\dist},
    classifier/.style      = {line width=\lw, trapezium, trapezium angle=80, rotate=-90, minimum width=.35*\dist, minimum height=.15*\dist, draw, trapezium stretches=true, draw=white},
    decoder/.style      = {line width=\lw, trapezium, trapezium angle=80, rotate=90,  minimum width=.55*\dist, minimum height=.15*\dist, draw=white, trapezium stretches=true},
    mult/.style={path picture={
      \draw[black]
    (path picture bounding box.south east) -- (path picture bounding box.north west) (path picture bounding box.south west) -- (path picture bounding box.north east);
    }},
    add/.style={path picture={
      \draw[black]
    (path picture bounding box.south) -- (path picture bounding box.north) (path picture bounding box.west) -- (path picture bounding box.east);
    }}
}

\makeatletter
\pgfplotsset{
  tufte axes/.style =
    {
      after end axis/.code =
        {
          \draw ({rel axis cs:0,0} -| {axis cs:\pgfplots@data@xmin,0})
            -- ({rel axis cs:0,0}  -| {axis cs:\pgfplots@data@xmax,0});
          \draw ({rel axis cs:0,0} |- {axis cs:0,\pgfplots@data@ymin})
            -- ({rel axis cs:0,0}  |-{axis cs:0,\pgfplots@data@ymax});
        },
      axis line style = {draw = none},
      tick align      = outside,
      tick pos        = left
    }
}
\makeatother

\pgfkeys{/pgfplots/tuftelike/.style={
  thick,
  tick style={major tick length=3pt, thick, black},
  axis x line*=bottom,
  axis line shift = 8pt,
  tick align      = outside,
  tick pos        = left,
  xlabel shift=0pt,
  axis y line*=left,
  label style={font=\small},
  tick label style={font=\small},
  ylabel shift=0pt}
}

\newcommand{\dist}{10em}

\usetikzlibrary{bayesnet}

\pgfplotsset{
    discard if not/.style 2 args={
        x filter/.code={
            \edef\tempa{\thisrow{#1}}
            \edef\tempb{#2}
            \ifx\tempa\tempb
            \else
                
            \fi
        }
    }
}

\pgfplotsset{
legend image code/.code={
\draw[mark repeat=2,mark phase=2]
plot coordinates {
(0cm,0cm)
(0.15cm,0cm)        
(0.3cm,0cm)         
};%
}
}

\pgfmathsetlengthmacro\MajorTickLength{
  \pgfkeysvalueof{/pgfplots/major tick length} * 4
}

\pgfplotsset{
    legend image with text/.style={
        legend image code/.code={%
            \node[anchor=center] at (0.15cm,0cm) {#1};
        }
    },
}

\pgfkeys{/pgfplots/pairedscatter/.style={
    tuftelike,
    cycle list/Dark2,
    y tick label style={
        /pgf/number format/.cd,
            fixed,
            fixed zerofill,
            precision=1,
        /tikz/.cd
    },
    ymin=50,
    ymax=100,
    ytick distance=10,
    ymajorgrids=true,
    legend pos=north east,
    legend columns=2,
    legend style={
        draw = none, at={(1.55,1.0)}, font=\tiny,
        legend style={
            legend cell align=left,
        },
    }
}}

\usepackage{dsfont}

\usepackage{soul}

\definecolor{gray.220}{RGB}{220, 220, 220}

\usepackage{xcolor,colortbl}
\definecolor{Gray}{gray}{0.95}
\newcolumntype{a}{>{\columncolor{Gray}}c}
\newcolumntype{d}{>{\columncolor{Gray}}r}

\usepackage{colortbl}
\usepackage{booktabs} 

\usepackage{csvsimple}
\usepackage{siunitx}
\usepackage[font=small]{caption}
\usepackage{tabularx}
\usepackage[export]{adjustbox}

\usepackage{cite}

\newcommand{\iid}{\emph{i.i.d.}}

\usepackage{amsmath}
\usepackage{environ}
\NewEnviron{neuripsonly}{%
}

\usepackage{amsthm}
\newtheorem{prop}{Proposition}

\newtheorem{defi}{Definition}
\newtheorem{lemma}{Lemma}

\newcommand{\tablefont}{\footnotesize}

\newcommand{\ObjN}{ON}

\title{Improving robustness against common corruptions\\by covariate shift adaptation}

\newcommand*\samethanks[1][\value{footnote}]{\footnotemark[#1]}

\author{
Steffen Schneider\thanks{Equal contribution. $^\dagger$ Equal contribution.; Online version and code: \href{https://domainadaptation.org/batchnorm}{domainadaptation.org/batchnorm}}\\
University of Tübingen \&\\ IMPRS-IS\\
\And
Evgenia Rusak\samethanks\\
University of Tübingen \&\\ IMPRS-IS\\
\And
Luisa Eck \\
LMU Munich \\
\hphantom{University of Tübingen \&}
\AND
Oliver Bringmann$^\dagger$\\
University of Tübingen\\
\And
Wieland Brendel$^\dagger$\\
University of Tübingen\\
\And
Matthias Bethge$^\dagger$ \\
University of Tübingen \\ 
}

\begin{document}

\doparttoc
\faketableofcontents

\maketitle

\begin{abstract}
Today's state-of-the-art machine vision models are vulnerable to image corruptions like blurring or compression artefacts, limiting their performance in many real-world applications.
We here argue that popular benchmarks to measure model robustness against common corruptions (like ImageNet-C) underestimate model robustness in many (but not all) application scenarios.
The key insight is that in many scenarios, multiple unlabeled examples of the corruptions are available and can be used for unsupervised online adaptation.
Replacing the activation statistics estimated by batch normalization on the training set with the statistics of the corrupted images consistently improves the robustness across 25 different popular computer vision models.
Using the corrected statistics, ResNet-50 reaches 62.2\% mCE on ImageNet-C compared to 76.7\% without adaptation.
With the more robust DeepAugment+AugMix model, we improve the state of the art achieved by a ResNet50 model up to date from 53.6\% mCE to 45.4\% mCE.
Even adapting to a single sample improves robustness for the ResNet-50 and AugMix models, and 32 samples are sufficient to improve the current state of the art for a ResNet-50 architecture.
We argue that results with adapted statistics should be included whenever reporting scores in corruption benchmarks and other out-of-distribution generalization settings.
\end{abstract}

\section{Introduction}\label{sec:introduction}
Deep neural networks (DNNs) are known to perform well in the independent and identically distributed (\iid{}) setting when the test and training data are sampled from the same distribution.
However, for many applications this assumption does not hold.
In medical imaging, X-ray images or histology slides will differ from the training data if different acquisition systems are being used.
In quality assessment, the images might differ from the training data if lighting conditions change or if dirt particles accumulate on the camera.
Autonomous cars may face rare weather conditions like sandstorms or big hailstones.
While human vision is quite robust to those deviations~\citep{geirhos2018noise}, modern machine vision models are often sensitive to such image corruptions. 

We argue that current evaluations of model robustness underestimate performance in many (but not all) real-world scenarios.
So far, popular image corruption benchmarks like ImageNet-C [IN-C;~\citealp{hendrycks2018benchmarking}] focus only on ad hoc scenarios in which the tested model has zero prior knowledge about the corruptions it encounters during test time, even if it encounters the same corruption multiple times.
In the example of medical images or quality assurance, the image corruptions do not change from sample to sample but are continuously present over a potentially large number of samples.
Similarly, autonomous cars will face the same weather condition over a continuous stream of inputs during the same sand- or hailstorm.
These (unlabeled) observations can allow recognition models to adapt to the change in the input distribution.

Such unsupervised adaptation mechanisms are studied in the field of domain adaptation (DA), which is concerned with adapting models trained on one domain (the source, here clean images) to another for which only unlabeled samples exist (the target, here the corrupted images).
Tools and methods from domain adaptation are thus directly applicable to increase model robustness against common corruptions, but so far no results on popular benchmarks have been reported.
The overall goal of this work is to encourage stronger interactions
between the currently disjoint fields of domain adaptation and robustness towards common corruptions.

We here focus on one popular technique in DA, namely adapting batch normalization [BN;~\citealp{ioffe2015batch}] statistics~\citep{schneider2018multi, cariucci2017autodial, DBLP:journals/corr/LiWSLH16}.
In computer vision, BN is a popular technique for speeding up training and is present in almost all current state-of-the-art image recognition models.
BN estimates the statistics of activations for the training dataset and uses them to normalize intermediate activations in the network.

By design, activation statistics obtained during training time do not reflect the statistics of the test distribution when testing in out-of-distribution settings like corrupted images.
We investigate and corroborate the hypothesis that high-level distributional shifts from clean to corrupted images largely manifest themselves in a difference of first and second order moments in the internal representations of a deep network, which can be mitigated by adapting BN statistics, i.e. by estimating the BN statistics on the corrupted images.
We demonstrate that this simple adaptation alone can greatly increase recognition performance on corrupted images.

Our contributions can be summarized as follows:
\begin{itemize}
    \item We suggest to augment current benchmarks for common corruptions with two additional performance metrics that measure robustness after partial and full unsupervised adaptation to the corrupted images.
    \item We draw connections to domain adaptation and show that even adapting to a single corrupted sample improves the baseline performance of a ResNet-50 model trained on IN from \num{76.7}\% mCE to \num{71.4}\%.
    Robustness increases with more samples for adaptation and converges to a mCE of \num{62.2}\%. 
    \item We show that the robustness of a variety of vanilla models trained on ImageNet [IN;~\citealp{russakovsky2015imagenet, deng2009imagenet}] substantially increases after adaptation, sometimes approaching the current state-of-the-art performance on IN-C without adaptation.
    \item Similarly, we show that the robustness of state-of-the-art ResNet-50 models on IN-C consistently increases when adapted statistics are used.
    We surpass the best non-adapted model (\num{52.3}\% mCE) by almost  \num{7}\% points.
    \item We show results on several popular image datasets and discuss both the generality and limitations of our approach.
    \item We demonstrate that the performance degradation of a non-adapted model can be well predicted from the Wasserstein distance between the source and target statistics. We propose a simple theoretical model for bounding the Wasserstein distance based on the adaptation parameters. 
\end{itemize}


\section{Measuring robustness against common corruptions}

The ImageNet-C benchmark~\citep{hendrycks2018benchmarking} consists of 15 test corruptions and four hold-out corruptions which are applied with five different severity levels to the \num{50000} test images of the  \textsc{ilsvrc2012} subset of ImageNet~\citep{deng2009imagenet}.
During evaluation, model responses are assumed to be conditioned only on single samples, and are not allowed to adapt to e.g. a batch of samples from the same corruption.
We call this the ad hoc or non-adaptive scenario.
The main performance metric on IN-C is the mean corruption error (mCE), which is obtained by normalizing the model's top-1 errors with the top-1 errors of AlexNet~\citep{krizhevsky2012imagenet} across the $C = 15$ test corruptions and $S = 5$ severities (cf.~\citealp{hendrycks2018benchmarking}):
\begin{equation}
    \text{mCE(model)} = \frac{1}{C} \sum_{c = 1}^{C}
    \frac{\sum_{s=1}^{S} \text{err}^\text{model}_{c,s}}{\sum_{s=1}^{S} \text{err}_{c,s}^\text{AlexNet}}.
\end{equation}
Note that mCE reflects only one possible averaging scheme over the IN-C corruption types.
We additionally report the overall top-1 accuracies and report results for all individual corruptions in the supplementary material and the project repository.

In many application scenarios, this ad hoc evaluation is too restrictive.
Instead, often many unlabeled samples with similar corruptions are available, which can allow models to adapt to the shifted data distribution.
To reflect such scenarios, we propose to also benchmark the robustness of adapted models.
To this end, we split the \num{50000} validation samples with the same corruption and severity into batches with $n$ samples each and allow the model to condition its responses on the complete batch of images.
We then compute mCE and top-1 accuracy in the usual way.

We consider three scenarios:
In the \emph{ad hoc} scenario, we set $n = 1$ which is the typically considered setting.
In the \emph{full adaptation} scenario, we set $n = \num{50000}$, meaning the model may adapt to the full set of unlabeled samples with the same corruption type before evaluation.
In the \emph{partial adaptation} scenario, we set $n = 8$ to test how efficiently models can adapt to a relatively small number of unlabeled samples.

\section{Correcting Batch Normalization statistics as a strong baseline for reducing covariate shift induced by common corruptions}\label{sec:methods}

We propose to use a well-known tool from domain adaptation---adapting batch normalization statistics~\citep{cariucci2017autodial,DBLP:journals/corr/LiWSLH16}---as a simple baseline to increase robustness against image corruptions in the adaptive evaluation scenarios.
IN trained models typically make use of batch normalization [BN;~\citealp{ioffe2015batch}] for faster convergence and improved stability during training.
Within a BN layer, first and second order statistics $\mu_c, \sigma^2_c$ of the activation tensors $\zz_c$ are estimated across the spatial dimensions and samples for each feature map $c$.
The activations are then normalized by subtracting the mean $\mu_c$ and dividing by $\sigma^2_c$.
During training, $\mu_c$ and $\sigma^2_c$ are estimated \emph{per batch}.
During evaluation, $\mu_c$ and $\sigma^2_c$ are estimated \emph{over the whole training dataset}, typically using exponential averaging~\citep{paszke2017automatic}.

Using the BN statistics obtained during training for testing makes the model decisions deterministic but is also problematic if the input distribution changes.
If the activation statistics $\mu_c, \sigma^2_c$ change for samples from the test domain, then the activations of feature map $c$ are no longer normalized to zero mean and unit variance, breaking a crucial assumption that all downstream layers depend on.
Mathematically, this \emph{covariate shift}\footnote{Note that our notion of internal covariate shift differs from previous work~\citep{ioffe2015batch,santurkar2018does}:
In \iid{} training settings,~\citet{ioffe2015batch} hypothesized that covariate shift introduced by changing lower layers in the network is reduced by BN, explaining the empirical success of the method.
We do not provide evidence for this line of research in this work: Instead, we focus on the covariate shift introduced (by design) in datasets such as IN-C, and provide evidence for the hypothesis that high-level domain shifts in the input partly manifests in shifts and scaling of \emph{internal} activations.}
can be formalized as follows:

\begin{defi}[Covariate Shift, cf.~\citealp{sugiyama2012machine,schoelkopf2012causal}]
There exists covariate shift between a source distribution with density $p_s:  \mathcal X \times \mathcal Y \rightarrow \mathbb{R}^+$ and a target distribution with density $p_t: \mathcal X \times \mathcal Y \rightarrow \mathbb{R}^+$, written as $p_s(\xx,y) = p_s(\xx) p_s(y|\xx)$ and $p_t(\xx,y) = p_t(\xx) p_t(y|\xx)$, if $p_s(y|\xx) = p_t(y|\xx)$ and $p_s(\xx) \neq p_t(\xx)$
where $y \in \mathcal{Y}$ denotes the class label.
\end{defi}

\paragraph{Removal of covariate shift.}
If covariate shift (Def. 1) only 
causes differences in the first and second order moments of the feature activations $\zz = f(\xx)$, it can be removed by applying normalization:
\begin{equation}
    p\left(\frac{f(\xx) - \Exp_s[f(\xx)]}{\sqrt{\mathbb{V}_s[f(\xx)]}}  \Big| \xx \right) p_s(\xx) \approx
    p\left(\frac{f(\xx) - \Exp_t[f(\xx)]}{\sqrt{\mathbb{V}_t[f(\xx)]}}  \Big| \xx \right) p_t(\xx).
\end{equation}

Reducing the covariate shift in models with batch normalization is particularly straightforward: it suffices to estimate the BN statistics $\mu_t, \sigma^2_t$ on (unlabeled) samples from the test data available for adaptation.
If the number of available samples $n$ is too small, the estimated statistics would be too unreliable.
We therefore leverage the statistics $\mu_s, \sigma^2_s$ already computed on the training dataset as a prior and infer the test statistics for each test batch as follows,
\begin{equation}
\Bar{\mu} = \frac{N}{N + n} \mu_s + \frac{n}{N + n} \mu_t,  \quad
\Bar{\sigma}^2 = \frac{N}{N + n} \sigma_s^2 + \frac{n}{N + n} \sigma_t^2.
\label{eq:Nn}
\end{equation}
The hyperparameter $N$ controls the trade-off between source and estimated target statistics and has the intuitive interpretation of a \emph{pseudo sample size} (p.~117,~\citealp{bishop}) for samples from the training set.
The case $N \to \infty$ ignores the test set statistics and is equivalent to the standard ad hoc scenario while $N = 0$ ignores the training statistics.
Supported by empirical and theoretical results (see results section and appendix), we suggest using $N\in[8,128]$ for practical applications with small $n < 32$.

\section{Experimental Setup}\label{sec:experiments}

\paragraph{Models.}%
We consider a large range of models (cf. Table~2, \textsection\ref{app:experimental-setup},\ref{apx:imagenet_model_descriptions}) 
and evaluate pre-trained variants of
DenseNet~\citep{densenet},
GoogLeNet~\citep{googlenet},
Inception and GoogLeNet~\citep{inception},
MNASnet~\citep{mnasnet},
MobileNet~\citep{mobilenet},
ResNet~\citep{he2016resnet},
ResNeXt~\citep{xie2017aggregated},
ShuffleNet~\citep{shufflenet},
VGG~\citep{vgg}
and Wide Residual Network [WRN,~\citealp{wideresnet}] from the \texttt{torchvision} library~\citep{10.1145/1873951.1874254}.
All models are trained on the \textsc{ilsvrc2012} subset of IN comprised of 1.2 million images in the training and a total of \num{1000} classes~\citep{russakovsky2015imagenet, deng2009imagenet}.
We also consider a ResNeXt-101 variant pre-trained on a 3.5 billion image dataset and then fine-tuned on the IN training set~\citep{mahajan2018exploring}.
We evaluate 3 models from the SimCLRv2 framework~\citep{chen2020big}.
We additionally evaluate the four leading methods from the ImageNet-C leaderboard, namely Stylized ImageNet training [SIN;~\citealp{geirhos2018imagenettrained}],
adversarial noise training [ANT;~\citealp{Rusak2020IncreasingTR}] as well as
a combination of ANT and SIN~\citep{Rusak2020IncreasingTR},
optimized data augmentation using AutoAugment [AugMix;~\citealp{hendrycks2019augmix,autoaugment2019}] and
Assemble Net~\citep{lee2020compounding}.
For partial adaptation, we choose $N \in \{2^{0},\cdots,2^{10}\}$ and select the optimal value on the holdout corruption mCE.

\paragraph{Datasets.}%
ImageNet-C [IN-C;~\citealp{hendrycks2018benchmarking}] is comprised of corrupted versions of the \num{50000} images in the IN validation set.
The dataset offers five severities per corruption type, for a total of 15 ``test'' and 4 ``holdout'' corruptions.
ImageNet-A [IN-A;~\citealp{DBLP:journals/corr/abs-1907-07174}] consists of unmodified real-world images which yield chance level classification performance in IN trained ResNet-50 models.
ImageNet-V2 [IN-V2;~\citealp{recht2019imagenet}] aims to mimic the test distribution of IN, with slight differences in image selection strategies.
ObjectNet [\ObjN{};~\citealp{barbu2019objectnet}] is a test set containing \num{50000} images like IN organized in 313 object classes with 109 unambiguously overlapping IN classes.
ImageNet-R [IN-R;~\citealp{hendrycks2020many}] contains \num{30000} images with various artistic renditions of 200 classes of the original IN dataset. 
Additional information on the used models and datasets can be found in \textsection\ref{app:experimental-setup}.
For IN, we resize all images to $256\times 256$px and take the center $224 \times 224$px crop.
For IN-C, images are already cropped.
We also center and re-scale the color values with $\mu_{RGB} = [0.485, 0.456, 0.406]$ and $\sigma=[0.229, 0.224, 0.225]$.

\section{Results}\label{sec:results}

\pgfplotsset{every tick label/.append style={font=\footnotesize}}
\pgfmathsetlengthmacro\MajorTickLength{
  \pgfkeysvalueof{/pgfplots/major tick length} * 4
}
\pgfkeys{%
    /pgfplots/plotone/.style={
        mark=star,
        mark options={solid,fill=none}
    },
    /pgfplots/plotsecond/.style={
        mark=triangle,
        mark options={solid,fill=none},
    },
    /pgfplots/plottwo/.style={
        mark=none
    },
    /pgfplots/plotthree/.style={
        mark=none
    },
    /pgfplots/plotfour/.style={
        mark=none
    }
}

\pgfplotsset{cycle list/Dark2-8}
\pgfplotsset{cycle list/Dark2-4}

\begin{figure}[tpb]
\begin{center}
\begin{tikzpicture}
\begin{axis}[
tuftelike,
y tick label style={
    /pgf/number format/.cd,
        fixed,
        fixed zerofill,
        precision=1,
    /tikz/.cd
},
width=0.4\textwidth,
height=0.33\textwidth,
cycle multi list={%
        Dark2-8\nextlist
        solid, dashed, dashdotted
},
xlabel=Batch size,
ylabel={mCE},
xmode=log,
xtick=      {1,2,4,8,16,32,64,128,256,512,1024,2048,50000},
xticklabels={1, , ,8,  ,  ,64,   ,   ,512,    ,    ,50\,000},
ytick={40, 60, 80, 100, 120, 140},
yticklabels={40, 60, 80, 100, 120, 140},
xmin=1,
xmax=50000,
ymin=40,
ymax=140,
legend pos=north east,
legend columns=2,
transpose legend,
legend style={draw = none, at={(1.2,0.99), font=\scriptsize}, name=legend},
name=ax1
]

\pgfplotsset{cycle list shift=0} 
\addplot+ [densely dashed, mark=none] table [x=batchsize, y=mCE]{data/batchsize/inc-rn50.tsv}; \label{pl:rn50-0}
\addplot+ [solid, mark=none] table [x=batchsize, y=resnet]{data/batchsize/rn50-augmix-opt.tsv}; \label{pl:rn50-16}
\addplot+ [plottwo] coordinates { (1, 76.7) (50000, 76.7) }; \label{pl:rn50-infty}

\addplot+ [densely dashed, mark=none] table [x=batchsize, y=mCE]{data/batchsize/inc-augmix.tsv}; \label{pl:am-0}
\addplot+ [solid, mark=none] table [x=batchsize, y=augmix]{data/batchsize/rn50-augmix-opt.tsv}; \label{pl:am-16}
\addplot+ [plotthree] coordinates { (1, 65.27) (50000, 65.27) }; \label{pl:am-infty}

\addplot+ [plotfour, draw = black!80] coordinates { (1, 53.6) (50000, 53.6) }; \label{pl:sota}

\addplot [mark=star, draw = none] coordinates { (8,48.4) (4096,45.4) }; \label{pl:sota-adapt}

\end{axis}
\node[draw=none,inner sep=2pt,below right] at (1, 3.25) {%
\scriptsize
\setlength{\tabcolsep}{2pt}
\begin{tabular}{ccl}
RN50 & AM \\
\ref{pl:rn50-infty} & \ref{pl:am-infty} & $N=\infty$ (base) \\
\ref{pl:rn50-0}     & \ref{pl:am-0} & $N=0$ (ours)\\
\ref{pl:rn50-16}    & \ref{pl:am-16} & $N$ best (ours)\\[.25em]
\multicolumn{3}{l}{DAug+AM (RN-50 SoTA)} \\
\ref{pl:sota} &  & $N = \infty$ (base)\\
\ref{pl:sota-adapt} &  & $N$ best (ours)
\end{tabular}};

\pgfplotsset{cycle list/Paired}

\begin{axis}[pairedscatter,
width=0.40\textwidth,
height=0.33\textwidth,
at={(0.5\textwidth,0)},
xlabel={IN top1 error},
ylabel={IN-C mCE},
xmin=20,
xmax=35,
ymin=50,
ymax=100,
cycle multi list={%
        Paired
},
name=ax2
]

\addplot+ [%
    draw = black!90,
    fill = black!90,
    opacity=.8,
	mark size = 0.15em,
    mark options = {solid},
	scatter/classes={%
        True={mark=*},
        False={mark=o}%
    },
	scatter,
	scatter src=explicit symbolic,
	discard if not={modelclass}{resnet}%
] table [
    x=in,
    y=inc,
    meta=trainmode
]{data/imagenet-c/torchvision-reordered.tsv};

\pgfplotsset{cycle list shift=-1} 

\foreach \model in {%
densenet,
resnext,
wide,
mnasnet,
mobilenet,
shufflenet,
googlenet,
inception,
vgg,
} {
\addplot+ [%
	mark size = 0.15em,
    opacity=.95,
    mark options = {solid},
	scatter/classes={%
        True={mark=*},
        False={mark=o}%
    },
	scatter,
	scatter src=explicit symbolic,
	discard if not={modelclass}{\model}%
] table [
    x=in,
    y=inc,
    meta=trainmode
]{data/imagenet-c/torchvision-reordered.tsv};
}
\legend{%
{ }, ResNet,
{ }, DenseNet,
{ }, ResNeXt,
{ }, WRN,
{ }, MNASnet,
{ }, MobileNet,
{ }, ShuffleNet,
{ }, GoogLeNet,
{ }, Inception,
{ }, VGG,
}

\end{axis}


\begin{scope}[every node/.style={text width=6.5cm,align=left}]
\node [below = 1.25 of ax1.south] {\captionof{figure}{Sample size vs. performance tradeoff in terms of the mean corruption error (mCE) on IN-C for ResNet-50 and AugMix (AM). Black line corresponds to (non-adapted) ResNet50 state-of-the-art performance of DeepAug+AugMix.}
\label{fig:low-resource}};
\node [below = 1.25 of ax2.south] {\captionof{figure}{Across 25 model architectures in the \texttt{torchvision} library, the baseline mCE ($\circ$) improves with adaptation ($\bullet$), often on the order of 10 points. Best viewed in color.}
\label{fig:torchvision}};

\end{scope}

\end{tikzpicture}

\end{center}
\end{figure}

\paragraph{Adaptation boosts robustness of a vanilla trained ResNet-50 model.}\label{sec:result-vanilla}

We consider the pre-trained ResNet-50 architecture from the \texttt{torchvision} library and adapt the running mean and variance on all corruptions and severities of IN-C for different batch sizes. 
The results are displayed in Fig.~\ref{fig:low-resource} where different line styles of the green lines show the number of pseudo-samples $N$ indicating the influence of the prior given by the training statistics.
With $N=16$, we see that even adapting to a single sample can suffice to increase robustness, suggesting that even the ad hoc evaluation scenario can benefit from adaptation.
If the training statistics are not used as a prior ($N=0$), then it takes around 8 samples to surpass the performance of the non-adapted baseline model (76.7\% mCE).
After around 16 to 32 samples, the performance quickly converges to 62.2\% mCE, considerably improving the baseline result.
These results highlight the practical applicability of batch norm adaptation in basically all application scenarios, independent of the number of available test samples.

\paragraph{Adaptation consistently improves corruption robustness across IN trained models.}
\label{sec:result-torchvision}

To evaluate the interaction between architecture and BN adaptation, we evaluate all 25 pre-trained models in the \texttt{torchvision} package and visualize the results in Fig.~\ref{fig:torchvision}.
All models are evaluated with $N=0$ and $n=\num{2000}$.
We group models into different families based on their architecture and observe consistent improvements in mCE for all of these families, typically on the order of 10\% points.
We observe that in both evaluation modes, DenseNets~\citep{densenet} exhibit higher corruption robustness despite having a comparable or even smaller number of trainable parameters than ResNets which are usually considered as the relevant baseline architecture.
A take-away from this study is thus that model architecture alone plays a significant role for corruption robustness and the ResNet architecture might not be the optimal choice for practical applications.

\paragraph{Adaptation yields new state of the art on IN-C for robust models.}\label{sec:result-sota}

\newcommand{\insrules}{%
        \cmidrule( r){1-1}
        \cmidrule(lr){2-5}
        \cmidrule(l ){6-9}
}

\begin{table}[tpb]

    \centering
    \footnotesize
    \caption{%
    Adaptation improves mCE (lower is better) and Top1 accuracy (higher is better) on IN-C for different models and surpasses the previous state of the art without adaptation. We consider $n=8$ for partial adaptation.
    }
\setlength{\aboverulesep}{0pt}
\setlength{\belowrulesep}{0pt}
      \begin{tabular}{l  c   c a  d c c a d }
         \toprule
          & \multicolumn{4}{c}{IN-C mCE ($\searrow$)}
          &  \multicolumn{4}{c}{Top1 accuracy ($\nearrow$)}
          \\
          & w/o &  partial & full & &  w/o &  partial & full & \\
          Model & adapt & adapt &  adapt & \multicolumn{1}{>{\columncolor{Gray}}c}{$\Delta$}  & adapt &  adapt&   adapt  & \multicolumn{1}{>{\columncolor{Gray}}c}{$\Delta$}  \\
         \insrules
         Vanilla ResNet-50 
         & \num{76.69} & \num{65.0151} & \num{62.24} & (\num{-14.45}) 
         & \num{39.17} & \num{48.5868} & \num{50.70} & (+\num{11.53})  \\
         \insrules
         SIN \citep{geirhos2018imagenettrained}
         & \num{69.32} & \num{61.4528} & \num{59.49} & (\num{-9.83}) 
         & \num{45.22} & \num{51.5846} & \num{53.11} & (+\num{7.89})\\
         ANT \citep{Rusak2020IncreasingTR} 
         & \num{63.37} & \num{56.1009} & \num{53.57} & (\num{-9.80}) 
         & \num{50.43} & \num{56.1107} & \num{58.00} & (+\num{7.57})\\
         ANT+SIN \citep{Rusak2020IncreasingTR} 
         & \num{60.66} & \num{55.3004} & \num{53.64} & (\num{-7.02})
         & \num{52.6}  & \num{56.8241} & \num{57.97} & (+\num{5.37})\\
         AugMix [AM; \citealp{hendrycks2019augmix}]
         & \num{65.27} & \num{55.4046} & \num{51.0} & (\num{-14.26})
         & \num{48.34} & \num{56.2959} &  \num{59.8} & (+\num{11.41})\\
         Assemble Net \citep{lee2020compounding}
         & \num{52.26} & -- & \num{50.10} & (\num{-1.16})
         & \num{59.22} & -- & \num{60.75} & (+\num{1.53})\\
         DeepAug \citep{hendrycks2020many} 
         &  \num{60.36} & \num{52.3282} &  \num{49.44} & (\num{-10.92})
         & \num{52.6}     & \num{59.012} & \num{61.2}    & (+\num{8.6})\\
         DeepAug+AM  \citep{hendrycks2020many}
         & \num{53.55}     & \num{48.3561} &  \num{45.36} & (\num{-8.19})
         & \num{58.1}     & \num{62.1973} & \num{64.531066}    &  (+\num{6.4})\\
         \cmidrule( r){1-9}
         DeepAug+AM+RNXt101 \citep{hendrycks2020many} & \textbf{44.5} & \num{40.67453} & \textbf{38.0} & (\num{-6.56}) 
         & \textbf{65.2} & \num{68.2319} &  \textbf{70.3} & (+\num{5.1}) \\
         \bottomrule
    \end{tabular}
    \label{tbl:sota}
\end{table}




We now investigate if BN adaptation also improves the most robust models on IN-C.
The results are displayed in Table~\ref{tbl:sota}.
All models are adapted using $n = \num{50000}$ (vanilla) or $n=\num{4096}$ (all other models) and $N=0$.
The performance of all models is considerably higher whenever the BN statistics are adapted.
The DeepAugment+AugMix reaches a new state of the art on IN-C for a ResNet-50 architecture of 45.4\% mCE.
Evaluating the performance of AugMix over the number of samples for adaptation (Fig.~\ref{fig:low-resource}, we find that as little as eight samples are sufficient to improve over AssembleNet~\citep{lee2020compounding}, the current state-of-the-art ResNet-50 model on IN-C without adaptation.
We have included additional results in \textsection\ref{app:results}.

\section{Analysis and Ablation Studies}
\label{ref:analysis}

\paragraph{Severity of covariate shift correlates with performance degradation.}

\pgfkeys{/pgfplots/wasserstein-scatter/.style={
    tuftelike,
    width=0.3\textwidth,
    height=0.3\textwidth,
    xlabel={avg $W$ (across layers)},
    ylabel shift = -.75 em,
    xlabel shift = -.55 em,
    label style={font=\tiny},
    ymax=100,
    ymin=20,
    ytick distance=20,
    xmin=0.0,
    xmax=4.0,
    yticklabel style = {font=\tiny},
    xticklabel style = {font=\tiny},
    cycle multi list={%
        Dark2\nextlist
        mark=star,mark=triangle,mark=square,mark=o,mark=x
    },
    legend columns=4,
    legend style={draw = none, at={(1.5,-.4), anchor=west}, name=legend, font=\tiny}
}}

\newcommand{\corruptionlist}{%
gaussian-noise, shot-noise, impulse-noise,{},speckle-noise,
defocus-blur, glass-blur, motion-blur, zoom-blur, gaussian-blur,
snow, frost, fog, brightness, spatter,
contrast, elastic-transform, pixelate, jpeg, saturate%
}

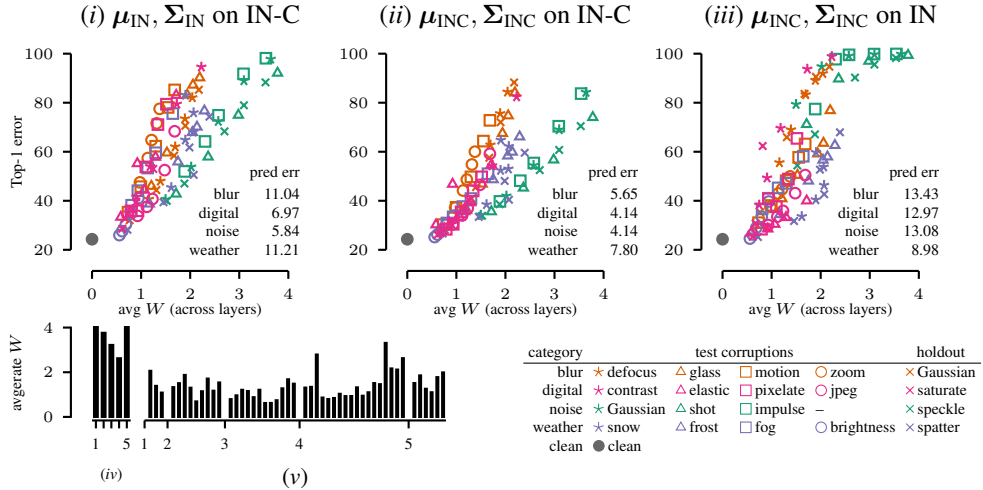
\begin{figure}[tpb]
\begin{center}
\begin{tikzpicture}
\pgfplotsset{cycle list/Dark2} 

\begin{axis}[wasserstein-scatter,
ylabel={Top-1 error},
title={(\emph{i}) $\mumu_\text{IN},\covd_\text{IN}$ on IN-C}]
\foreach \corruption in \corruptionlist {
\addplot+ [
    only marks,
    discard if not={corruption}{\corruption}%
    ] table [
    x=wasserstein, 
    y=error] {data/analysis/wasserstein/vanilla_on_inc.tsv};
    \label{pl:\corruption}%
};
\addplot [mark=*,black!60, only marks] coordinates { (0, 24.314) };
\label{pl:baseline}

\end{axis}

\node[draw=none,inner sep=2pt] at (2,.5) {%
\tiny
\setlength{\tabcolsep}{2pt}
\begin{tabular}{rr}
        & \multicolumn{1}{c}{pred err} \\[.25em]
blur    &        11.04 \\
digital &         6.97 \\
noise   &         5.84 \\
weather &        11.21 \\
\end{tabular}
};

\node[draw=none,inner sep=2pt] at (6.5,.5) {%
\tiny
\setlength{\tabcolsep}{2pt}
\begin{tabular}{rr}
        & \multicolumn{1}{c}{pred err} \\[.25em]
blur    &         5.65 \\
digital &         4.14 \\
noise   &         4.14 \\
weather &         7.80 \\
\end{tabular}
};

\node[draw=none,inner sep=2pt] at (10.5,.5) {%
\tiny
\setlength{\tabcolsep}{2pt}
\begin{tabular}{rr}
        & \multicolumn{1}{c}{pred err} \\[.25em]
blur    &        13.43 \\
digital &        12.97 \\
noise   &        13.08 \\
weather &         8.98 \\
\end{tabular}
};

\node[draw=none,inner sep=2pt] at (8.75,-2) {%
\tiny
\setlength{\tabcolsep}{2pt}
\begin{tabular}{rlllll}
category & \multicolumn{4}{c}{test corruptions} & \multicolumn{1}{c}{holdout} \\
\hline
blur
&\ref{pl:defocus-blur} defocus
&\ref{pl:glass-blur}     glass
&\ref{pl:motion-blur}  motion
&\ref{pl:zoom-blur}  zoom
&\ref{pl:gaussian-blur}  Gaussian \\
digital
&\ref{pl:contrast} contrast
&\ref{pl:elastic-transform}     elastic
&\ref{pl:pixelate}  pixelate
&\ref{pl:jpeg}  jpeg
&\ref{pl:saturate} saturate \\
noise
&\ref{pl:gaussian-noise}  Gaussian
&\ref{pl:shot-noise}  shot
&\ref{pl:impulse-noise} impulse 
& --
&\ref{pl:speckle-noise}  speckle \\
weather
&\ref{pl:snow} snow
&\ref{pl:frost}     frost 
&\ref{pl:fog}  fog 
&\ref{pl:brightness}  brightness
&\ref{pl:spatter} spatter\\
clean
& \multicolumn{2}{l}{\ref{pl:baseline}  clean} \\
\end{tabular}
};

\begin{axis}[wasserstein-scatter,
title={(\emph{ii}) $\mumu_\text{INC},\covd_\text{INC}$ on IN-C},
at={(.3\textwidth,0)}]
\foreach \corruption in \corruptionlist {,
\addplot+ [
    only marks,
    discard if not={corruption}{\corruption}%
    ] table [
    x=wasserstein, 
    y=error] {data/analysis/wasserstein/adapt_on_inc.tsv};
}
\addplot [mark=*,black!60, only marks] coordinates { (0, 24.314) };
\end{axis}

\begin{axis}[wasserstein-scatter,
title={(\emph{iii}) $\mumu_\text{INC},\covd_\text{INC}$ on IN},
at={(.6\textwidth,0)}]
\foreach \corruption in \corruptionlist {,
\addplot+ [
    only marks,
    discard if not={corruption}{\corruption}%
    ] table [
    x=wasserstein, 
    y=error] {data/analysis/wasserstein/adapt_on_in.tsv};
}
\addplot [mark=*,black!60, only marks] coordinates { (0, 24.314) };
\end{axis}

\begin{axis}[%
    tuftelike,
    ybar,
    bar width=0.05cm,
    cycle list/Dark2,
    axis x line shift = 1pt,
    at={(0,-.16\textwidth)},
    width=0.15\textwidth,
    yticklabel style = {font=\tiny},
    xticklabel style = {font=\tiny},
    label style={font=\tiny},
    enlargelimits=0.01,
    height=0.2\textwidth,
    xlabel={(\emph{iv})},
    ylabel={avgerate $W$},
    xmin=-0.5, xmax=4.5,
    ymin=0, ymax=4,
    ytick = {0, 2, 4},
    enlarge x limits=-0.5,
    xtick = {0,1,2,3,4},
    xticklabels = {1,,,,5}
]
\addplot+ [fill=black,draw=black] table [
    y=difference] {data/analysis/wasserstein-layer-norm-downsample.tsv};
\end{axis}

\begin{axis}[%
    tuftelike,
    ybar,
    bar width=0.025cm,
    cycle list/Dark2,
    y axis line style={draw=none},
    axis x line shift = 1pt,
    yticklabel style = {font=\tiny},
    xticklabel style = {font=\tiny},
    ytick = {},
    at={(.05\textwidth,-.16\textwidth)},
    width=0.4\textwidth,
    height=0.2\textwidth,
    xlabel={(\emph{v})},
    ymajorticks=false,
    xmin=1, xmax=53.5,
    xtick = {1,5,15,28,47},
    xticklabels = {1,2,3,4,5},
    ymin=0, ymax=4
]
\addplot+ [fill=black!100,draw=black!100] table [
    y=difference] {data/analysis/wasserstein-layer-norm.tsv};
\end{axis}
\end{tikzpicture}
\end{center}

\caption{%
The Wasserstein metric between optimal source (IN) and target (IN-C) statistics correlates well with top-1 errors
(\emph{i}) of non-adapted models on IN-C,
(\emph{ii}) of adapted models on IN-C, indicating that even after reducing covariate shift, the metric is predictive of the remaining source--target mismatch
(\emph{iii}) IN-C adapted models on IN, the reverse case of (\emph{i}).
Holdout corruptions can be used to get a linear estimate on the prediction error of test corruptions (tables).
We depict input and downsample (\emph{iv}) as well as bottlneck layers (\emph{v}) and notice the largest shift in early and late downsampling layers.
The metric is either averaged across layers (\emph{i}--\emph{iii}) or across corruptions (\emph{iv}--\emph{v}).
\label{fig:wasserstein-analysis-detailed}
}
\end{figure}

The relationship between the performance degradation on IN-C and the covariate shift suggests an unsupervised way of estimating the classification performance of a model on a new corruption.
Taking the normalized Wasserstein distance (cf. \textsection\ref{app:distance}) between the statistics of the source and target domains\footnote{For computing the Wasserstein metric we make the simplifying assumption that the empirical mean and covariances fully parametrize the respective distributions.} computed on all samples with the same corruption and severity and averaged across all network layers, we find a correlation with the top-1 error (Fig.~\ref{fig:wasserstein-analysis-detailed} \emph{i--iii}) of both non-adapted (\emph{i}) and fully adapted model (\emph{ii}) on IN-C corruptions.
Within single corruption categories (noise, blur, weather, and digital), the relationship between top-1 error and Wasserstein distance is particularly striking: using linear regression, the top-1 accuracy of hold-out corruptions can be estimated with around 1--2\% absolute mean deviation (cf. \textsection\ref{app:error_prediction}) within a corruption, and with around 5--15\% absolute mean deviation when the estimate is computed on the holdout corruption of each category (see Fig.~\ref{fig:wasserstein-analysis-detailed}, typically, a systematic offset remains).
In Fig.~\ref{fig:wasserstein-analysis-detailed}\emph{(iv--v}), we display the Wasserstein distance across individual layers and observe that the covariate shift is particularly present in early and late downsampling layers of the ResNet-50.

\paragraph{Large scale pre-training alleviates the need for adaptation.}
\label{sec:ablation-wsl}

Computer vision models based on the ResNeXt architecture~\citep{xie2017aggregated} pretrained on a much larger dataset comprised of \num{3.5e9} Instagram images (IG-3.5B) achieve a 45.7\% mCE on IN-C~\citep{mahajan2018exploring,orhan2019robustness}.
We re-evaluate these models with our proposed paradigm and summarize the results in Table~\ref{tbl:wsl}.
While we see improvements for the small model pre-trained on IN, these improvements vanish once the model is trained on the full IG-3.5B dataset.
This observation also holds for the largest model, suggesting that training on very large datasets might alleviate the need for covariate shift adaptation.

\begin{table}[tpb]\footnotesize
    \begin{minipage}{.5\linewidth}
      \centering
      \setlength{\tabcolsep}{5pt}
      \captionsetup{width=.75\textwidth}
        \caption{Improvements from adapting the BN parameters vanish for models trained with weakly supervised pre-training.}

      \tablefont
          \begin{tabular}{l  c  c }
        \toprule
          & \multicolumn{2}{c}{IN-C mCE ($\searrow$)}
          \\
          ResNeXt101 & BN & BN+adapt  \\
        \cmidrule( r){1-1}
        \cmidrule(lr){2-3}
        32x8d, IN
        & \num{66.60} 
        & \num{56.70} (\num{-9.90}) \\
        32x8d, IG-3.5B
        & \num{51.70} 
        & \num{51.64} (\num{-0.06}) \\
        32x48d, IG-3.5B
        & \textbf{45.7}
        & \textbf{47.3} (+\num{1.56})\\
        \bottomrule
            \end{tabular}
      \label{tbl:wsl}
    \end{minipage}%
    \begin{minipage}{.5\linewidth}
      \centering
      
           \setlength{\tabcolsep}{5pt}
    \captionsetup{width=.95\textwidth}
    \caption{Fixup and GN trained models perform better than non-adapted BN models but worse than adapted BN models.}

      \tablefont
       \begin{tabular}{lcccc}
              \toprule
      & \multicolumn{4}{c}{IN-C mCE ($\searrow$)} \\
      Model    & Fixup & GN & BN & BN+adapt \\
      \midrule
      ResNet-50  & \num{72.00}   & \num{72.35} & \num{76.7}   & \textbf{62.2} \\
      ResNet-101 & \num{68.16}   & \num{67.62} & \num{69.01}  & \textbf{59.1} \\
      ResNet-152 & \num{67.58}   & \num{65.44} & \num{69.27}  & \textbf{58.0} \\
      \bottomrule
    \end{tabular} 
      \label{tbl:fixup}
    \end{minipage} 
\end{table}

\paragraph{Group Normalization and Fixup Initialization performs better than non-adapted batch norm models, but worse than batch norm with covariate shift adaptation.}

So far, we considered image classification models with BN layers and concluded that using training dataset statistics in BN generally degrades model performance in out-of-distribution evaluation settings.
We now consider models trained without BN and study the impact on corruption robustness, similar to~\citet{galloway2019batch}.

First, using Fixup initialization \citep{zhang2019fixup} alleviates the need for BN layers.
We train a ResNet-50 model on IN for 100 epochs to obtain a top-1 error of \num{24.23}\% and top-5 error of \num{ 7.56}\% (compared to  27.6\% reported by~\citet{zhang2019fixup} with shorter training, and the \num{23.87}\% obtained by our ResNet-50 baseline trained with BN).
The model obtains an IN-C mCE of \num{72.00}\% compared to \num{76.7}\% mCE of the vanilla ResNet-50 model and \num{62.2}\% mCE of our adapted ResNet-50 model (cf. Table~\ref{tbl:fixup}).
Additionally, we train a ResNet-101 and a ResNet-152 with Fixup initialization with similar results.
Second, GroupNorm [GN;~\citealp{wu2018group}] has been proposed as a batch-size independent normalization technique.
We train a ResNet-50, a ResNet-101 and a ResNet-152 architecture for 100 epochs and evaluate them on IN-C and find results very similar to Fixup.

\paragraph{Results on other datasets: IN-A, IN-V2, ObjectNet, IN-R}

\pgfkeys{/pgfplots/tuftelike/.style={
  thick,
  cycle list = {solid, densely dashed, densely dotted, densely dash dot},
  tick style={major tick length=3pt, thick, black},
  axis x line*=bottom,
  axis line shift = 10pt,
  tick align      = outside,
  tick pos        = left,
  xlabel shift=-5pt,
  axis y line*=left,
  ylabel shift=-7pt}
}

\definecolor{colorone}{HTML}{1b9e77}
\definecolor{colortwo}{HTML}{d95f02}

\pgfplotsset{
legend image code/.code={
\draw[mark repeat=2,mark phase=2]
plot coordinates {
(0cm,0cm)
(0.1cm,0cm)
(0.3cm,0cm)
};
}
}

\pgfkeys{/pgfplots/imagenetablation/.style={
    y tick label style={
        /pgf/number format/.cd,
            fixed,
            fixed zerofill,
            precision=1,
        /tikz/.cd
    },
    title style={font=\small},
    width=0.35\columnwidth,
    height=0.35\textwidth,
    xlabel=Batch size,
    ylabel={Top1 Error},
    xtick=      {1,2,4,8,16,32,64,128,256,512,1024,2048,4096},
    xticklabels={1, , ,8,  ,  ,64,   ,   ,512,    ,    ,4096},
    ytick={20,40,60, 80, 100, 120, 140},
    yticklabels={,,,,,,},
    xmin=1,
    xmax=4096,
    ymin=20,
    ymax=100,
    yticklabel style = {font=\scriptsize},
    xticklabel style = {font=\scriptsize},
    ylabel style = {font=\scriptsize},
    xlabel style = {font=\scriptsize},
    unbounded coords=jump,
    legend pos=north east,
    legend style={draw = none, at={(1.0,0.99)}, name=legend, font=\scriptsize}}
}
  
\pgfmathsetlengthmacro\MajorTickLength{
  \pgfkeysvalueof{/pgfplots/major tick length} * 4
}

\begin{figure}[tpb]
\begin{center}
\begin{tikzpicture}
\begin{axis}[%
    tuftelike,
    imagenetablation,
    title={ImageNet V1 vs. V2},
    at={(0,0)},
    ylabel={},
    ylabel={Top1 Error},
    yticklabels={20,40,60, 80, 100, 120, 140},
    xmode=log,
    legend style={draw = none, at={(1.0,0.99)}, name=legend}
]

\addplot+ [mark=circle, draw=colortwo] table [x=batchsize, y=error]{data/batchsize/imagenet-rn50.tsv};
\foreach \dataset in {threshold,topimages, matched}  {
    \addplot+ [
        mark=none,
        draw=colorone,
        discard if not={dataset}{\dataset}%
        ] table [x=batchsize, y=error]{data/batchsize/imagenet-v2-rn50.tsv};
}

\pgfplotsset{cycle list shift=-4}
\addplot+ [mark=none] coordinates { (1, 23.870) (4000, 23.870) };
\addplot+ [mark=none] coordinates { (1, 27.5524) (4000, 27.5524) }; 
\addplot+ [mark=none] coordinates { (1, 21.8581) (4000, 21.8581) }; 
\addplot+ [mark=none] coordinates { (1, 37.84) (4000, 37.84) }; 

\legend{%
    ImageNet,
    Threshold,
    Top,
    Matched,
}

\end{axis}
~
\begin{axis}[%
    tuftelike,
    imagenetablation,
    title={ImageNet vs. ObjectNet},
    ylabel={},
    at={(0.33\columnwidth,0)},
    legend style={at={(1.1,0.65)}, cells={align=left}},
    xmode=log,
]

\addplot+ [mark=circle, draw=colortwo] table [x=batchsize, y=error]{data/batchsize/imagenet-rn50.tsv};
\addplot+ [mark=none, draw=colorone] table [x=batchsize, y=error]{data/batchsize/objectnet-rn50.tsv};

\pgfplotsset{cycle list shift=-2} 
\addplot+ [mark=none] coordinates { (1, 23.870) (4000, 23.870) };
\addplot+ [mark=none] coordinates { (1, 78.147268) (4000, 78.147268) };

\legend{%
{ImageNet\\(1k classes)}, 
{ObjectNet\\(109 classes)}
}

\end{axis}
~
\begin{axis}[%
    tuftelike,
    imagenetablation,
    title={ImageNet-R},
    ylabel={},
    at={(0.66\columnwidth,0)},
    xmode=log,
    legend style={at={(1.0,0.3)}},
    legend columns = 2,
    legend cell align={left}
]

\addplot+ [%
    draw=colorone,
    discard if not={model}{baseline}%
] table [x=bsz, y=top1]{data/imagenet-r/batchsize.tsv};
\addplot+ [
    draw=colorone,
    discard if not={model}{augmix}%
] table [x=bsz, y=top1]{data/imagenet-r/batchsize.tsv};
\addplot+ [
    draw=colorone,
    discard if not={model}{deepaugmix}%
] table [x=bsz, y=top1]{data/imagenet-r/batchsize.tsv};

\pgfplotsset{cycle list shift=-3}
\addplot+ [mark=none] coordinates { (1, 63.8) (2048, 63.8) };
\addplot+ [mark=none] coordinates { (1, 59.0) (2048, 59.0) };
\addplot+ [mark=none] coordinates { (1, 53.2) (2048, 53.2) };

\legend{%
Baseline, AM, Deepaug+AM
}
\end{axis}
~

\end{tikzpicture}
\caption{\small%
Batch size vs. performance trade-off for different natural image datasets with no covariate shift (IN, IN-V2),
complex and shuffled covariate shift (ObjectNet),
complex and systematic covariate shift (ImageNet-R).
Straight black lines show baseline performance (no adaptation). ImageNet plotted for reference.
}
\label{fig:imagenet-v2}
\end{center}
\end{figure}
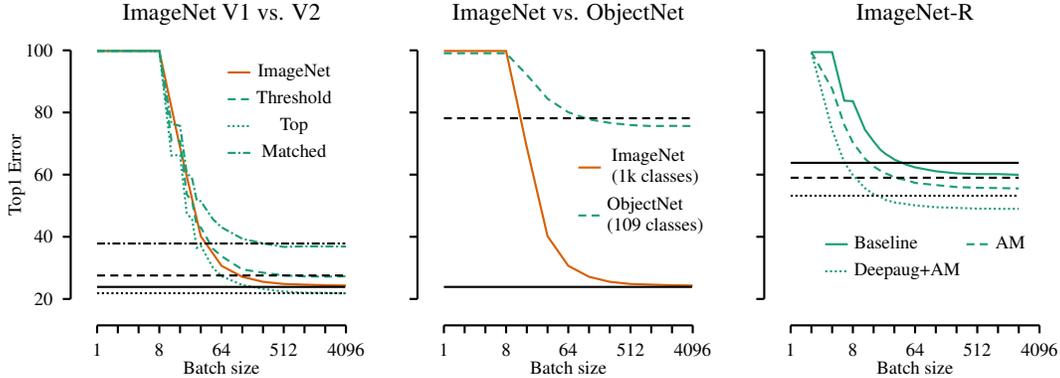

\newcommand{\insrulesINRTablesecond}{%
        \cmidrule( r){1-1}
        \cmidrule(lr){2-4}
}

\begin{table}[tpb]\footnotesize
    \begin{minipage}[t]{.5\linewidth}
      \centering
 \setlength{\tabcolsep}{5pt}
        \captionsetup{width=.9\textwidth, skip=14.7pt}
        \caption{GN and Fixup achieve the best results on ObjectNet (ON). After shuffling IN-C corruptions, BN adaptation does no longer decrease the error. Adaptation improves the performance of a vanilla ResNet50 on IN-R.
        }
\tablefont
\begin{tabular}{lccccc}
\toprule
& \multicolumn{2}{c}{ON} & \multicolumn{2}{c}{Mixed IN-C} & IN-R\\
ResNet50 & top-1 & top-5 & top-1 & top-5 & top-1\\
\midrule
BN w/o adapt & \num{78.15} & \num{60.91} & \num{61.08} &  \num{40.81} & \num{63.8} \\
BN w/ adapt & \num{75.96} & \num{58.85}  & \num{60.87} &  \num{40.31} & \textbf{59.9} \\
GroupNorm & \textbf{70.8} & \textbf{49.8} & \num{57.25} &  \num{35.97} & 61.2  \\
Fixup & \num{71.48} & \num{51.44}  & \textbf{56.8} &  \textbf{35.4} & 65.0 \\
\bottomrule
\end{tabular}
      \label{tbl:objectnet}
    \end{minipage}%
    \begin{minipage}[t]{.5\linewidth}
      \centering
    \small
    \captionsetup{width=.9\textwidth}
    \caption{%
Adaptation improves the performance (top-1 error) of robust models on IN-R (n=2048).
    }
\tablefont
\begin{tabular}{lrcr}
\toprule
Model & base & adapt & $\Delta$ \\
\insrulesINRTablesecond
ResNet50                   &  63.8 &   59.9 & -3.9 \\
SIN                        &  58.6 &   54.2 & -4.4 \\
ANT                        &  61.0 &   58.0 & -3.0 \\
ANT+SIN                    &  53.8 &   52.0 & -1.8 \\
AugMix (AM)                &  59.0 &   55.8 & -3.2 \\
DeepAug (DAug)              &  57.8 &   52.5 & -5.3\\
DAug+AM           &  53.2 &   48.9 & -4.3 \\
\insrulesINRTablesecond
DAug+AM+RNXt101 &  \textbf{47.9} &   \textbf{44.0} & -3.9\\
\bottomrule
\end{tabular}
    \label{tbl:imagenet-r}
    \end{minipage} 
\end{table}

We use $N=0$ and vary $n$ in all ablation studies in this subsection. 
The technique does not work for the case of ``natural adversarial examples'' of IN-A~\citep{DBLP:journals/corr/abs-1907-07174} and the error rate stays above 99\%, suggesting that the
covariate shift introduced in IN-A by design is more severe compared to the covariate shift of IN-C and can not be corrected by merely calculating the correct BN statistics.
We are not able to increase performance neither on IN nor on IN-V2, since in these datasets, no domain shift is present by design (see Fig.~\ref{fig:imagenet-v2}).
For ON, the performance increases slightly when computing statistics on more than 64 samples.
In Table~\ref{tbl:objectnet} (first and second column), we observe that the GroupNorm and Fixup models perform better than our BN adaptation scheme:
while there is a dataset shift in ON compared to IN, BN adaptation is only helpful for \textit{systematic} shifts across multiple inputs and this assumption is violated on ON.
As a control experiment, we sample a dataset ``Mixed IN-C'' where we shuffle the corruptions and severities.
In Table~\ref{tbl:objectnet} (third and fourth column), we now observe that BN adaptation expectedly no longer improves performance.
On IN-R, we achieve better results for the adapted model compared to the non-adapted model as well as the GroupNorm and Fixup models, see Table~\ref{tbl:objectnet} (last column). Additionally, on IN-R, we decrease the top-1 error for a wide range of models through adaptation (see Table~\ref{tbl:imagenet-r}).
For IN-R, we observe performance improvements for the vanilla trained ResNet50 when using a sample size of larger than 32 samples for calculating the statistics (Fig.~\ref{fig:imagenet-v2}, right-most plot).

\paragraph{A model for correcting covariate shift effects.}
\begin{wrapfigure}[14]{r}{0.3\textwidth}
  \begin{center}
\includegraphics[width=0.3\textwidth]{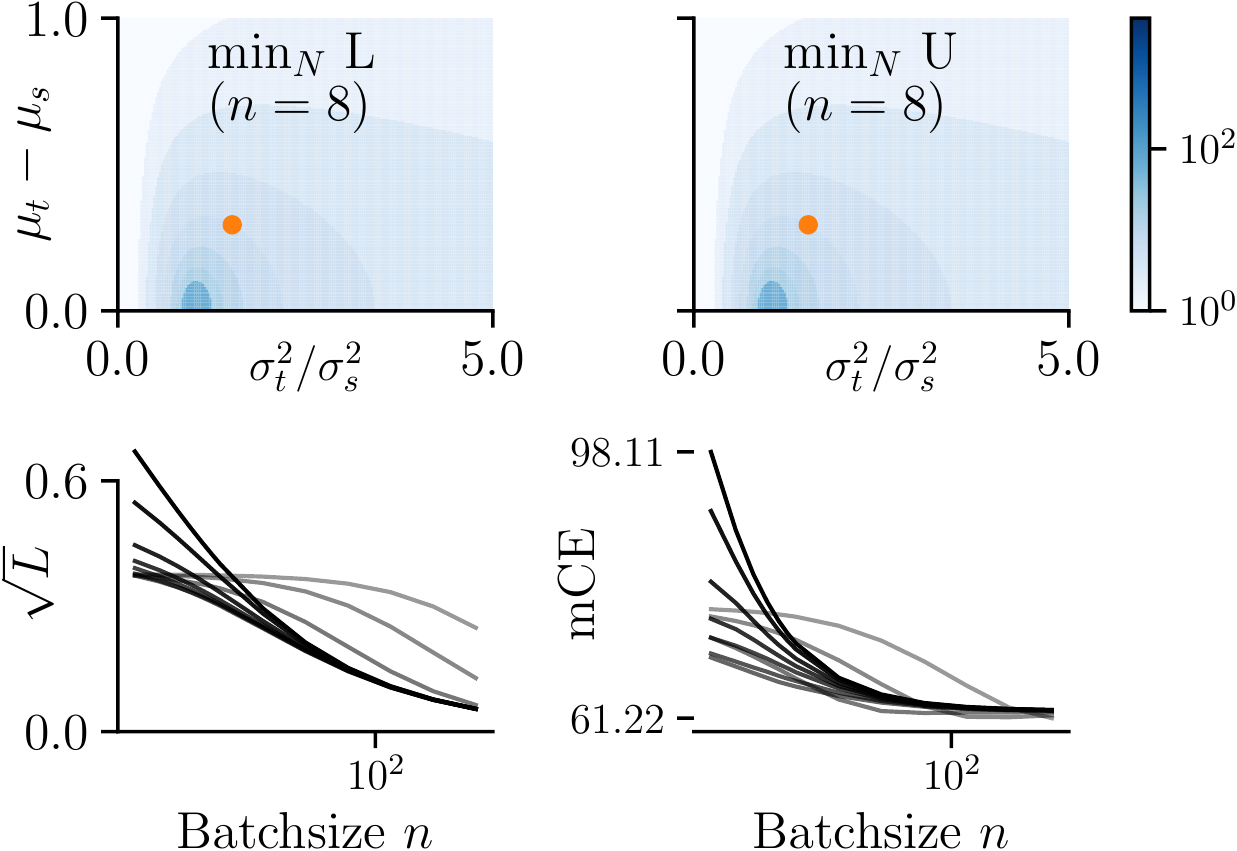}
 \end{center}
  \caption{%
   The bound suggests small optimal $N$ for most parameters (i) and qualitatively explains our empirical observation (ii).
   \label{eq:theory}
}
\end{wrapfigure}

We evaluate how the batch size for estimating the statistics at test time affects the performance on IN, IN-V2, \ObjN{} and IN-R in Fig.~\ref{fig:imagenet-v2}. 
As expected, for IN the adaptation to test time statistics converges to the performance of the train time statistics in the limit of large batch sizes, see Fig.~\ref{fig:imagenet-v2} middle.
For IN-V2, we find similar results, see Fig.~\ref{fig:imagenet-v2} left.
This observation shows that (\textit{i}) there is no systematic covariate shift between the IN train set and the IN-V2 validation set that could be corrected by using the correct statistics and (\textit{ii}) is further evidence for the \iid{} setting pursued by the authors of IN-V2.
In case of \ObjN{} (Fig.~\ref{fig:imagenet-v2} right), we see slight improvements when using a batch size bigger than 128. 

Choosing the number of pseudo-samples $N$ offers an intuitive trade-off between estimating accurate target statistics (low $N$) and relying on the source statistics (large $N$).
We propose a simple model to investigate optimal choices for $N$, disregarding all special structure of DNNs, and focusing on the statistical error introduced by estimating $\hat{\mu}_t$ and $\hat{\sigma}^2_t$ from a limited number of samples $n$.
To this end, we estimate upper ($U$) and lower ($L$) bounds of the expected squared Wasserstein distance $W_2^2$ as a function of $N$ and the covariate shift which provides good empirical fits between the estimated $W$ and empirical performance for ResNet-50 for different $N$ (Fig.~\ref{eq:theory}; bottom row).
Choosing $N$ such that $L$ or $U$ are minimized (Fig.~\ref{eq:theory}; example in top row) qualitatively matches the values we find, see \textsection\ref{app:analytical} for all details.

\begin{prop}[Bounds on the expected value of the Wasserstein distance between target and combined estimated target and source statistics]

We denote the source statistics as $\mu_s,\sigma_s^2$, the true target statistics as $\mu_t,\sigma^2_t$ and the biased estimates of the target statistics as $\hat{\mu}_t,\hat{\sigma}_t^2$. 
For normalization, we take a convex combination of the source statistics and estimated target statistics as discussed in Eq.~\ref{eq:Nn}.
At a confidence level $1-\alpha$, the expectation value of the Wasserstein distance
$W_2^2(\Bar{\mu}, \Bar{\sigma}, \mu_t, \sigma_t)$ between ideal and estimated target statistics w.r.t. to the distribution of sample mean $\hat{\mu}_t$ and sample variance $\hat{\sigma}^2_t$ is bounded from above and below
with $L \leq \mathds{E}[W^2_2] \leq U$, where 
\begin{equation*}
\begin{aligned}
    L &= \; \left( \sigma_t - \sqrt{ \frac{N}{N+n} \sigma^2_s + \frac{n-1}{N+n} \sigma^2_t } \right)^2 + \frac{N^2}{(N+n)^2} \left(\mu_t -  \mu_s \right)^2 + \frac{n}{(N+n)^2} \sigma_t^2 \\
    U &= \; L + \sigma^5_t \frac{(n-1)}{2(N+n)^2}  \left( \frac{N}{N+n} \sigma^2_s + \frac{1}{N+n} \chi^2_{1-\alpha/2, n-1} \sigma^2_t \right)^{-3/2}
\end{aligned}
\end{equation*}
The quantity $\chi^2_{1-\alpha/2, n-1}$ denotes the left tail value of a chi square distribution with $n-1$ degrees of freedom, defined as $P\left(X \leq \chi^2_{1-\alpha/2, n-1}\right) = \alpha/2 \text{ for } X \sim \chi^2_{n-1}$.
\emph{Proof:} See Appendix \textsection\ref{app:analytical}.
\end{prop}

\section{Related Work}\label{sec:related-work}

The IN-C benchmark~\citep{hendrycks2018benchmarking} has been extended to MNIST~\citep{mu2019mnist}, several object detection datasets~\citep{michaelis2019benchmarking} and image segmentation~\citep{Kamann2019BenchmarkingTR} reflecting the interest of the robustness community.
Most proposals for improving robustness involve special training protocols, requiring time and additional resources.
This includes data augmentation like Gaussian noise~\citep{ford2019adversarial}, optimized mixtures of data augmentations in conjunction with a consistency loss~\citep{hendrycks2019augmix}, training on stylized images~\citep{geirhos2018imagenettrained, michaelis2019benchmarking, Mikoajczyk2018DataAF} or against adversarial noise distributions~\citep{Rusak2020IncreasingTR}.
Other approaches tweak the architecture, e.g. by adding shift-equivariance with an anti-aliasing module,~\citep{zhang2019making} or assemble different training techniques~\citep{lee2020compounding}.

Unsupervised domain adaptation (DA) is a form of transductive inference where additional information about the test dataset is used to adapt a model to the test distribution.
Adapting feature statistics was proposed by~\citet{sun2017correlation} and follow up work evaluated the performance of adapting BN parameters in unsupervised \citep{DBLP:journals/corr/LiWSLH16, cariucci2017autodial} and supervised DA settings \citep{schneider2018multi}.
As an application example in medical imaging,~\citet{bug2017context} show that adaptive normalization is useful for removing domain shifts on histopathological data.
More involved methods for DA include self-supervised domain adaptation on single examples \citep{sun2019test} and pseudo-labeling \citet{french2017self}.
\citet{xie2020self} achieve the state of the art on IN-C with pseudo-labeling.
In work concurrent to ours, \citet{wang2020fully} also show BN adaptation results on IN-C. They also perform experiments on CIFAR10-C and CIFAR100-C and explore other domain adaptation techniques.

Robustness scores obtained by adversarial training can be improved when separate BN or GroupNorm layers are used for clean and adversarial images \citep{xie2020intriguing}.
The expressive power of adapting only affine BN parameters BN parameters was shown in multi-task \citep{rebuffi2017learning} and DA contexts \citep{schneider2018multi} and holds even for fine-tuning randomly initialized ResNets \citep{frankle2020training}.  
Concurrent work shows additional evidence that BN adaptation yields increased performance on ImageNet-C~\citep{nado2020evaluating}. 

\section{Discussion and Conclusion}\label{sec:discussion}

We showed that reducing covariate shift induced by common image corruptions improves the robustness of computer vision models trained with BN layers, typically by 10--15\% points (mCE) on IN-C.
Current state-of-the-art models on IN-C can benefit from adaptation, sometimes drastically like AugMix (\num{-14}\% points mCE).
This observation underlines that current benchmark results on IN-C underestimate the corruption robustness that can be reached in many application scenarios where additional (unlabeled) samples are available for adaptation.

Robustness against common corruptions improves even if models are adapted only to a single sample, suggesting that BN adaptation should always be used whenever we expect machine vision algorithms to encounter out-of-domain samples. Most further improvements can be reaped by adapting to 32 to 64 samples, after which additional improvements are minor.

Our empirical results suggest that the performance degradation on corrupted images can mostly be explained by the difference in feature-wise first and second order moments.
While this might sound trivial, the performance could also degrade because models mostly extract features susceptible to common corruptions~\citep{geirhos2020shortcut}, which could not be fixed without substantially adapting the model weights.
The fact that model robustness increases after correcting the BN statistics suggests that the features upon which the models rely on are still present in the corrupted images. The opposite is true in other out-of-domain datasets like IN-A or ObjectNet where our simple adaptation scheme does not substantially improve performance, suggesting that here the main problem is in the features that models have learned to use for prediction.

Batch Norm itself is not the reason why models are susceptible to common corruptions. While alternatives like Group Normalization and Fixup initialization slightly increase robustness, the adapted BN models are still substantially more robust.
This suggests that non-BN models still experience an internal covariate shift on corrupted images, but one that is now absorbed by the model parameters instead of being exposed in the BN layers, making it harder to fix.

Large-scale pre-training on orders of magnitude more data (like IG-3.5B) can remove the first- and second-order covariate shift between clean and corrupted image samples, at least partially explaining why models trained with weakly supervised training~\citep{mahajan2018exploring} generalize so well to IN-C.

Current corruption benchmarks emphasize ad hoc scenarios and thus focus and bias future research efforts on these constraints.
Unfortunately, the ad hoc scenario does not accurately reflect the information available in many machine vision applications like classifiers in medical computer vision or visual quality inspection algorithms, which typically encounter a similar corruption continuously and could benefit from adaptation.
This work is meant to spark more research in this direction by suggesting two suitable evaluation metrics---which we strongly suggest to include in all future evaluations on IN-C---as well as by highlighting the potential that even a fairly simple adaptation mechanism can have for increasing model robustness.
We envision future work to also adopt and evaluate more powerful domain adaptation methods on IN-C and to develop new adaptation methods specifically designed to increase robustness against common corruptions.

\section*{Broader Impact}

The primary goal of this paper is to increase the robustness of machine vision models against common corruptions and to spur further progress in this area. Increasing the robustness of machine vision systems can enhance their reliability and safety, which can potentially contribute to a large range of use cases including autonomous driving, manufacturing automation, surveillance systems, health care and others. Each of these uses may have a broad range of societal implications: autonomous driving can increase mobility of the elderly and enhance safety, but could also enable more autonomous weapon systems. Manufacturing automation can increase resource efficiency and reduce costs for goods, but may also increase societal tension through job losses or increase consumption and thus waste. Of particular concern (besides surveillance) is the use of generative vision models for spreading misinformation or for creating an information environment of uncertainty and mistrust.

We encourage further work to understand the limitations of machine vision models in out-of-distribution generalization settings. More robust models carry the potential risk of automation bias, i.e., an undue trust in vision models. However, even if models are robust to common corruptions, they might still quickly fail on slightly different perturbations like surface reflections. Understanding under what conditions model decisions can be deemed reliable or not is still an open research question that deserves further attention.

\begin{ack}

We thank
Julian Bitterwolf,
Roland S. Zimmermann,
Lukas Schott,
Mackenzie W. Mathis,
Alexander Mathis,
Asim Iqbal,
David Klindt,
Robert Geirhos,
other members of the Bethge and Mathis labs
and four anonymous reviewers
for helpful suggestions for improving our manuscript and providing ideas for additional ablation studies.
We thank the International Max Planck Research School for Intelligent Systems (IMPRS-IS) for supporting E.R. and St.S.;
St.S. acknowledges his membership in the European Laboratory for Learning and Intelligent Systems (ELLIS) PhD program.
This work was supported by the German Federal Ministry of Education and Research (BMBF) through the Tübingen AI Center (FKZ: 01IS18039A),
by the Deutsche Forschungsgemeinschaft (DFG) in the priority program 1835 under grant BR2321/5-2
and by SFB 1233, Robust Vision: Inference Principles and Neural Mechanisms (TP3), project number: 276693517.
The authors declare no conflicts of interests.

\end{ack}

\medskip
\small
\bibliographystyle{unsrtnat}
\bibliography{main}
\normalsize

\clearpage
\appendix
\normalsize
\part{Supplementary Material}
\parttoc
\clearpage
\section{Distances and divergences for quantifying domain shift}
\label{app:distance}

Besides analyzing the performance drop when evaluating a model using source statistics on a target dataset, we consider the mismatch in model statistics directly.
We first take an ImageNet trained model and adapt it to each of the 95 conditions in IN-C.
To obtain a more exact estimate of the true statistics, we split the model into multiple stages with only few BN layers per stage and apply the following simple algorithm\footnote{%
Note that for simplicity, we do not reset the statistics of the remaining $(b_i - i)$ BN layers. This could potentially be adapted in future work.
}:

\begin{itemize}
    \item Start with image inputs $\zz^0_n \leftarrow \xx_n$ from the validation set to adapt to, for each $n\in[50000]$.
    \item Split the model into multiple stages,
    $h(\xx) = (f_{m} \circ \cdots \circ f_1)(\xx)$,
    where each module $f_i$ can potentially contain one or multiple BN layers.
    We denote the number of BN layers in the $i$-th module as $b_i$.
    \item For each stage $i \in [m]$, repeat $b_i$ times:
    $\zz^i_n \leftarrow f_i(\zz^{i-1}_n)$ for each $n$, and update the BN statistics in module $f_i(\zz^{i-1}_n)$.
    \item Return $h$ with adapted statistics.
\end{itemize}

Using this scheme, we get source statistics $\mu_s$ and $\Sigma_s$ for each layer and $\mu_t$ and $\Sigma_t$ for each layer and corruption.
In total, we get 96 different collections of statistics across network layers (for IN and the 95 conditions in IN-C).
For simplicity, we will not further index the statistics.
Note that all covariance matrices considered here are diagonal, which is a further simplification.
We expect that our domain shift estimates could be improved by considering the full covariance matrices.

In the following, we will introduce three possible distances and divergences which can be applied between source and target statistics to quantify the effect of common corruptions induced covariate shift.
We consider the Wasserstein distance, a normalized version of the Wasserstein distance, and the Jeffrey divergence.

\subsection{The Wasserstein distance}
\label{def:wasserstein}

Given a baseline ResNet-50 model with source statistics $\mu_s, \Sigma_s$ on IN, the Wasserstein distance (cf. \citealp{villani2008optimal}) between the train and test distribution with statistics $\mu_t, \Sigma_t$ is given as 
\begin{equation}
W_2 (p_s, p_t)^2 = \| \mumu_s - \mumu_t \|_2^2 + \text{tr} \left( \cov_s + \cov_t - 2 \left(\cov_t^{1/2} \cov_s \cov_t^{1/2} \right)^{1/2}\right).
\end{equation}

\subsection{The source-normalized Wasserstein distance}
\label{def:wasserstein-norm}

When estimated for multiple layers across the network, the Wasserstein distance between source and target depends on the overall magnitude of the statistics.
Practically, this means the metric is dominated by features with large magnitude (e.g. in the first layer of a neural network, which receives larger inputs).

To mitigate this issue, we normalize both statistics with the source statistics and define the normalized Wasserstein distance as
\begin{align}
    \widetilde{W}_2^2 &= W_2^2\left(\covd_s^{-1/2} \mumu_s, \mathbf{I}, \covd_s^{-1/2} \mumu_t, \covd_s^{-1} \covd_t \right) \\
    &= \Tr{\left(\mathbf{I} + \covd_t \covd_s^{-1} - 2 \covd_t^{1/2} \covd_s^{-1/2} \right)} + (\mumu_t - \mumu_s)^T \covd_s^{-1} (\mumu_t - \mumu_s).
\end{align}

In the uni-variate case, the normalized Wasserstein distance $\widetilde{W}_2^2$ is equal to the Wasserstein distance $W_2^2$ between source and target statistics divided by $\sigma^2_s$:
\begin{align}
    \widetilde{W}_2^2 &= W_2^2\left(
    \frac{\mu_s}{\sigma_s},
    1, \frac{\mu_t}{\sigma_s},
    \frac{\sigma^2_t}{\sigma^2_s} \right) = 1 + \frac{\sigma^2_t}{\sigma^2_s} - 2 \frac{\sigma_t}{\sigma_s} + \frac{(\mu_t - \mu_s)^2}{\sigma^2_s} 
    = \frac{1}{\sigma^2_s} W_2^2(\mu_s, \sigma^2_s, \mu_t, \sigma^2_t).
\end{align}

\subsection{The Jeffrey divergence}

The Jeffrey divergence $J(p_s, p_t)$ between source distribution $p_s$ and target distribution $p_t$ is the symmetrized version of the Kullback-Leibler divergence $D_{KL}$:
\begin{align}
    J(p_s, p_t) = \frac{1}{2} \left( D_{KL}(p_s \| p_t) +  D_{KL}(p_t \| p_s) \right)
\end{align}
The Kullback-Leibler divergence between the $D$-dimensional multivariate normal source and target distributions is defined as
\begin{align}
    D_{KL}(\normaldist_t \| \normaldist_s) &= \frac{1}{2} \left( \Tr{\left(\cov_s^{-1} \cov_t\right)} + (\mumu_s - \mumu_t)^\top \cov_s^{-1} (\mumu_s - \mumu_t) - D + \ln{\left( \frac{\det \cov_s}{\det \cov_t}\right)} \right).
\end{align}
The Jeffrey divergence between the $D$-dimensional multivariate normal source and target distributions then follows as
\begin{align}
    J(\normaldist_t, \normaldist_s) &= \frac{1}{4} \left( \Tr{\left(\cov_s^{-1} \cov_t\right)} +  \Tr{\left(\cov_t^{-1} \cov_s\right)}  + (\mumu_s - \mumu_t)^\top \left(\cov_s^{-1} + \cov_t^{-1}\right) (\mumu_s - \mumu_t) - 2D \right).
\end{align}

\subsection{Summary statistics and quantification of covariate shift between different IN-C conditions}

\begin{figure}[hp!]
    \begin{center}
      \includegraphics[width=0.95\textwidth]{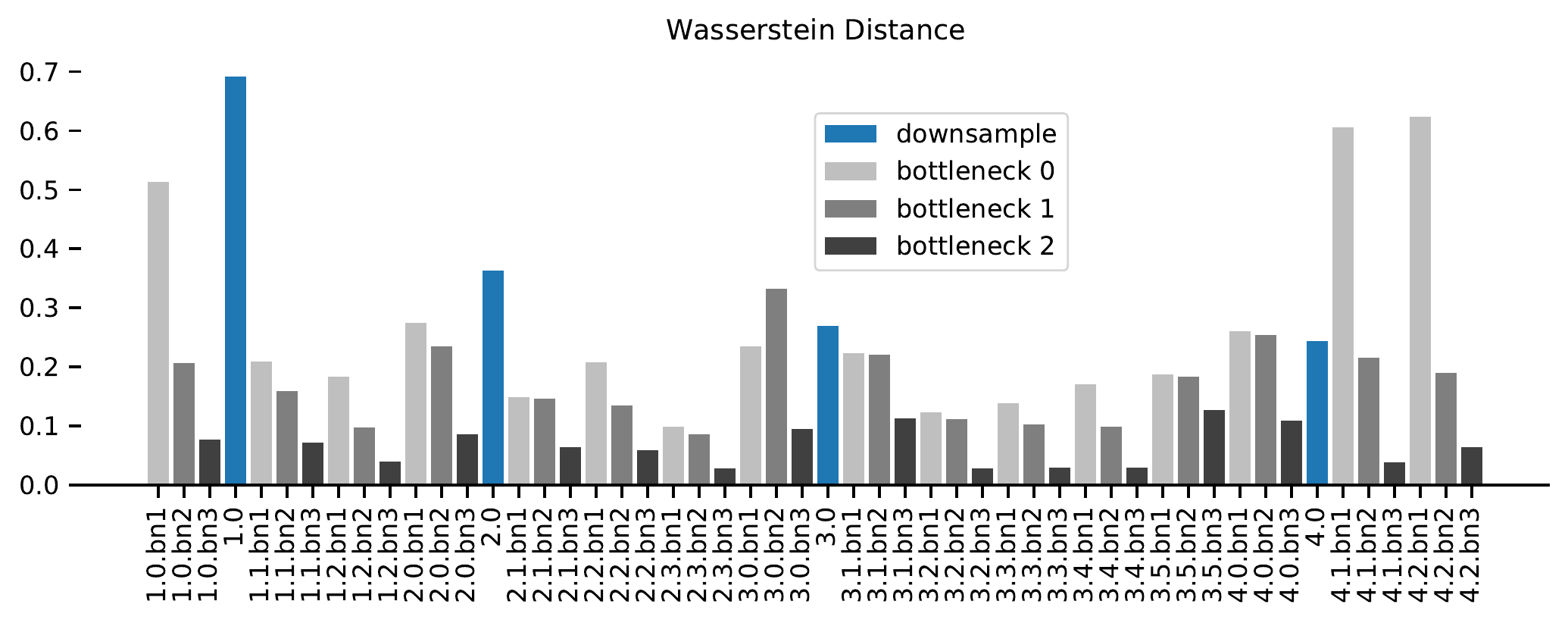}
      \end{center}
    \begin{center}
      \includegraphics[width=0.95\textwidth]{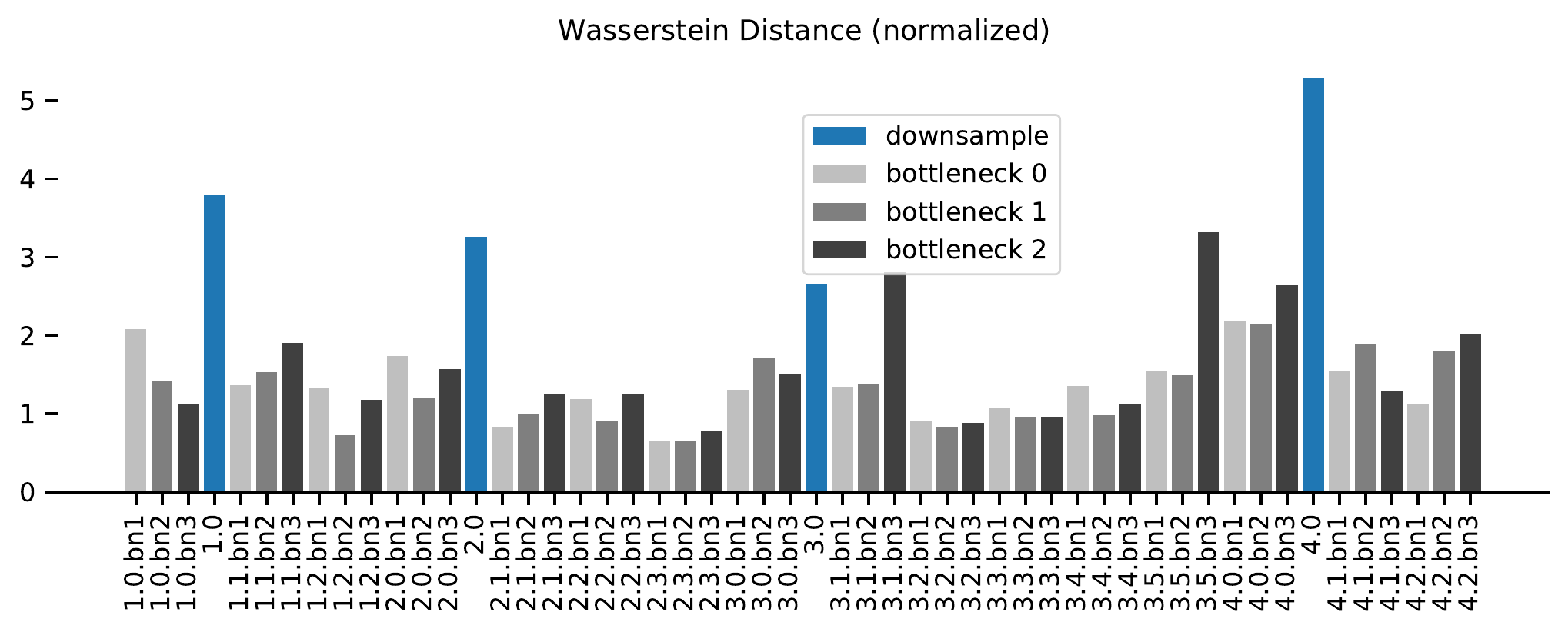}
      \end{center}
    \begin{center}
      \includegraphics[width=0.95\textwidth]{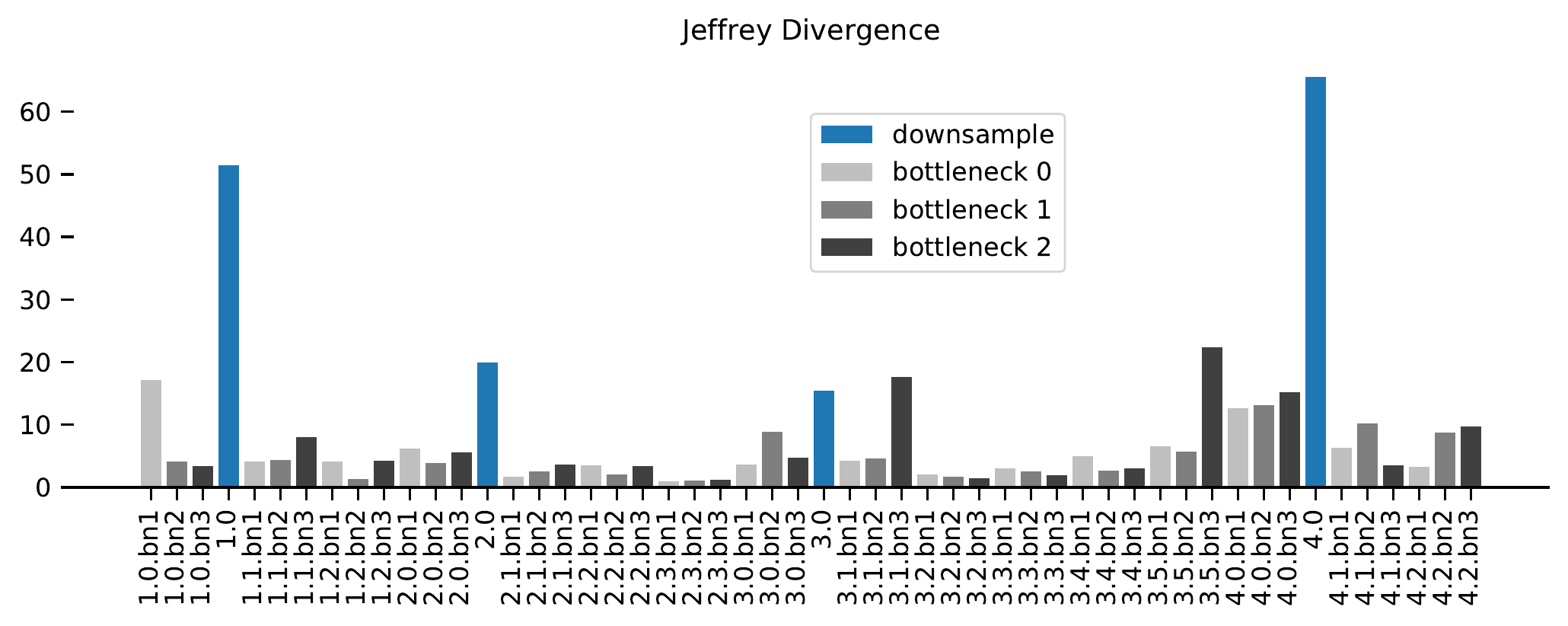}
      \end{center}
      \caption{
        Wasserstein distance,
        normalized Wasserstein distance and Jeffrey divergence estimated among source and target statistics between different network layers.
        We report the respective metric w.r.t. to the difference between baseline (IN) and target (IN-C) statistics and show the value averaged across all corruptions.
        We note that for a ResNet-50 model, downsampling layers contribute most to the overall error.
       }
       \label{fig:wasserstein-layers}
\end{figure}

Given the 95 distances/divergences between the baseline (IN) statistics and 95 IN-C conditions, we first perform a layer-wise analysis of the statistics and depict the results in Figure~\ref{fig:wasserstein-layers}.
The unnormalized Wasserstein distance is sensitive to the magnitude of the source statistics and hence differs qualitatively from the results on the normalized Wasserstein distance and Jeffrey Divergence.
We appreciate that the most notable difference between source and target domains is visible in the ResNet-50 downsampling layers.
All three metrics suggest that the shift is mainly present in the first and final layers of the network, supporting the hypothesis that within the common corruption dataset, we have both superficial covariate shift which can be corrected by simple means (such as brightness or contrast variations) in the first layers, and also more ``high-level'' domain shifts which can only be corrected in the later layers of the network.

\begin{figure}[hp!]
    \begin{center}
      \includegraphics[width=0.95\textwidth]{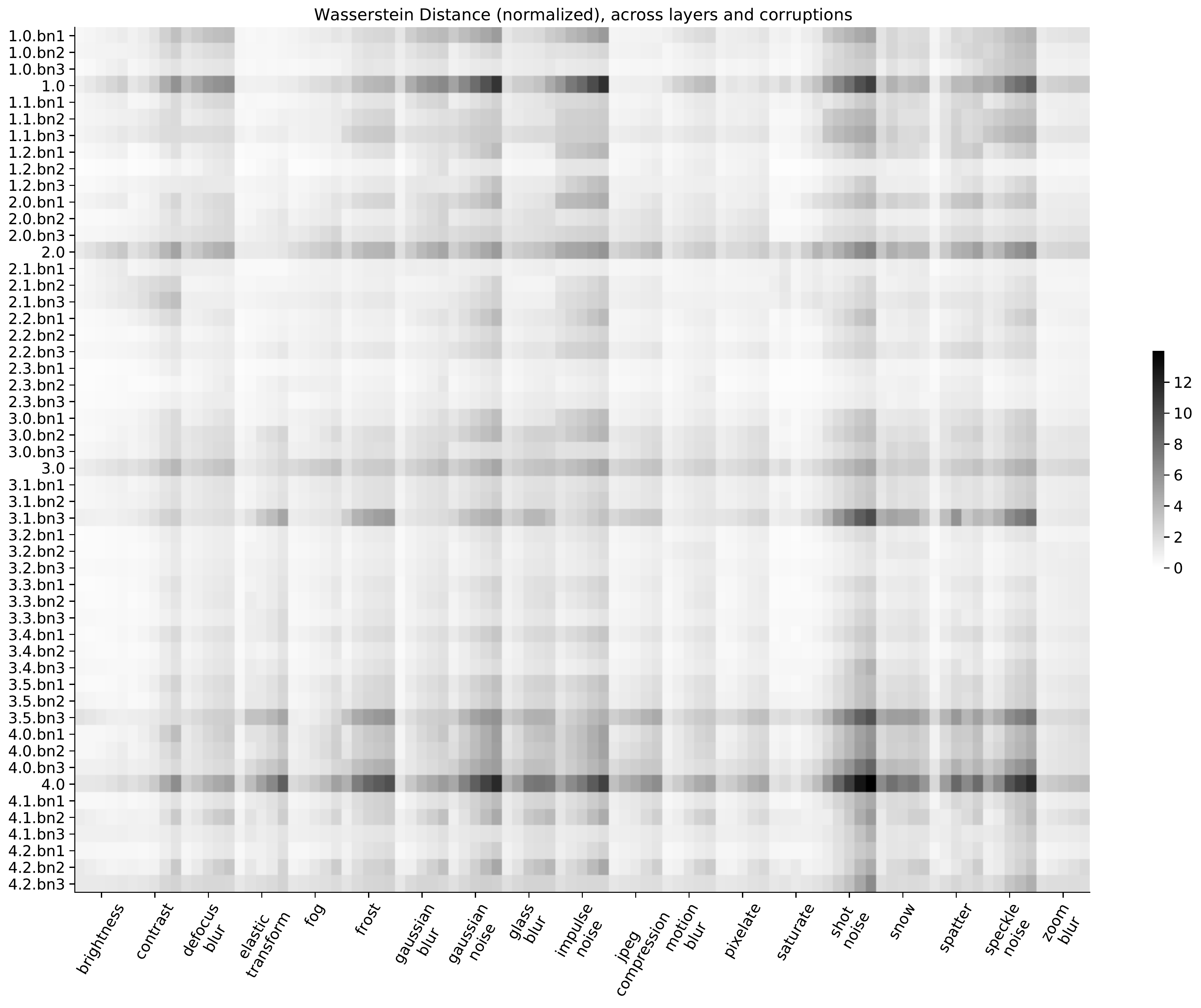}
      \end{center}
    \begin{center}
      \includegraphics[width=0.95\textwidth]{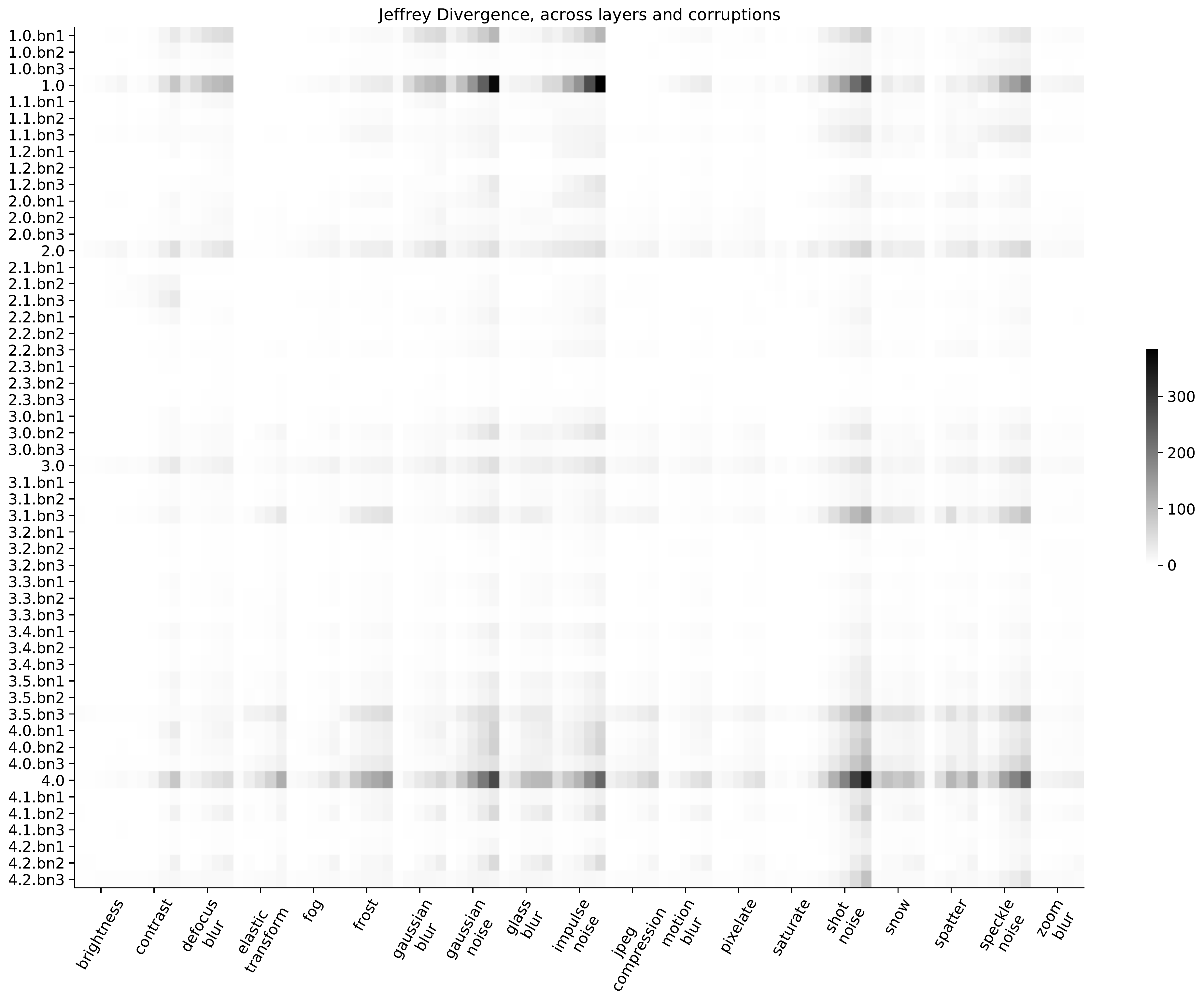}
      \end{center}
      \caption{
        Normalized Wasserstein distance and Jeffrey divergence across corruptions and layers in a ResNet-50.
       \label{fig:wasserstein-layers-fine}
       }
\end{figure}

In Figure~\ref{fig:wasserstein-layers-fine}, we more closely analyze this relationship for different common corruptions.
We can generally appreciate the increased measures as the corruption severity increases.

\clearpage
\section{Notes on the experimental setup}
\label{app:experimental-setup}

\subsection{Practical considerations for implementing the method}

Our method is conceptually very easy to implement.
We generally recommend to first explore the easier variant of the algorithm where $N=0$, i.e., no source statistics are used.
As shown in our experiments, this setting works well if 100 or more target samples are available.

In this case, implementing the method boils down to enabling the training mode for all BN layers across the network.
We will discuss this option along with two variants important for application to practical problems:
Using exponential moving averaging (EMA) to collect target statistics across multiple batches, and using the source statistics as a prior.

\paragraph{Example implementation in PyTorch and caveats}

We encourage authors of robust models to always evaluate their models, and in particular baseline algorithms on both the train and test set statistics.
Implementation in both PyTorch, Tensorflow and other machine learning libraries is straightforward and adds only minimal overhead.
For PyTorch, adaptation is possible by simply adding

\begin{verbatim}
def use_test_statistics(module):
  if isisinstance(module, nn._BatchNorm):
    module.train()
model.eval()
model.apply(use_test_statistics)
\end{verbatim}
before starting a model evaluation.
For the adaptation to a full dataset, we provide a reference implementation with the source code release of this paper.
Also, in contrast to the convention of not shuffling examples during test time, \emph{make sure to enable dataset shuffling also during test time} in order to compute the correct statistics marginalized over class assignment.

\paragraph{Exponential moving averaging}

In practice, it might be beneficial to keep track of samples already encountered and use a running mean and variance on the test set to normalize new samples.
We can confirm that this technique closely matches the full-dataset adaptation case even when evaluating with batch size 1 and is well suited for settings with less powerful hardware, or in general settings where access to the full batch of samples is not possible.
Variants of this technique include the adaptation of the decay factor to discard statistics of samples encountered in the past (e.g. when the data domain slowly drifts over time).


\subsection{Notes on models}

Note that we only re-evaluate existing model checkpoints, and hence do not perform any hyperparameter tuning or adaptations to model training except for selecting the pseudo batchsize $N$ for the source domain.
Depending on the batch size and the architecture, model evaluations are done on one to eight Nvidia RTX 2080 GPUs (i.e., using 12 to 96 GB of memory) or up to four Nvidia V100 GPUs (128 GB of memory).
Since we merely re-evaluate trained models, it is also possible to work on less powerful hardware with less memory.
In these cases, the aggregation of batch normalization statistics has to be done across several batches using a variant of EMA.

\subsection{Hyperparameter tuning}

Our method is generally parameter-free if only target statistics should be considered for normalization.
This approach is generally preferred for larger batch sizes $n$ and should also be adapted in practice when a sufficient amount of samples is available.
For tuning $N$, we consider the pre-defined holdout corruptions in IN-C, including speckle noise, saturation, Gaussian blur and spatter using a grid search across different values for $N$.

\subsection{Notes on datasets}

In the main paper, we have used several datasets and provide more relevant information here:

\paragraph{ImageNet-C (IN-C)} For the evaluation on IN-C, we use the JPEG compressed images from \href{https://github.com/hendrycks/robustness}{github.com/hendrycks/robustness} as is advised by the authors to ensure reproducibility.
We note that \citet{ford2019adversarial} report a decrease in performance when the compressed JPEG files are used as opposed to applying the corruptions directly in memory without compression artefacts.

\paragraph{ObjectNet (ON)} We find that there are 9 classes with multiple possible mappings from ON to IN (see the list in Table~\ref{ON:ambiguous}); we discard these classes in our evaluation. Models trained on IN experience a large performance drop on the order of 40--45\% when tested on \ObjN{}.
\ObjN{} is an interesting test case for unsupervised domain adaptation since IN and \ObjN{} are likely sampled from different distributions. ON intentionally shows objects from new viewpoints on new backgrounds.
\paragraph{ImageNet-V2 (IN-V2)} There are three test sets in IN-V2 that differ in \emph{selection frequencies} of the MTurk workers. 
The selection frequency is given by the fraction of MTurk workers who selected an image for its target class.
For the ``MatchedFrequency'' dataset, images were sampled according to the estimated selection frequency of sampling of the original IN validation dataset.
For the ``Threshold0.7'' variant of IN-V2, images were sampled with a selection frequency of at least 0.7.
The ``TopImages'' was sampled from images with the highest selection frequency.
Although all three test sets were sampled from the same Flickr candidate pool and were labeled correctly and selected by more than 70\% of MTurk workers, the model accuracies on these datasets vary by 14\%.
The authors observe a systematic accuracy drop when comparing model performance on the original IN validation set and IN-V2 and attribute it to the distribution gap between their datasets and the original IN dataset.
They quantify the distribution gap by how much the change from the original distribution to the new distribution affects the considered model.
Engstrom et al. analyze the creation process of IN-V2 and identify statistical bias resulting from noisy readings of the selection frequency statistic as a main source of dropping performance \citep{engstrom2020identifying}. After correcting the bias, \citep{engstrom2020identifying} find that the accuracy drop between IN and IN-V2 measures only 3.6\% ± 1.5\% of the original 11.7\% ± 1.0\%.


\begin{table}[ht]
    \centering
    \small
    \begin{tabular}{l l }
         \toprule
          ON class & IN classes \\
         \midrule
         wheel & wheel; paddlewheel, paddle wheel \\
         helmet & football helmet; crash helmet \\
         chair & barber chair; folding chair; rocking chair, rocker \\
         still\_camera & Polaroid camera, Polaroid Land camera; reflex camera \\
         alarm\_clock & analog clock; digital clock \\
         tie & bow tie, bow-tie, bowtie; Windsor tie \\
         pen & ballpoint, ballpoint pen, ballpen, Biro; quill, quill pen; fountain pen  \\
         bicycle & mountain bike, all-terrain bike, off-roader; bicycle-built-for-two, tandem bicycle, tandem \\
         skirt & hoopskirt, crinoline; miniskirt, mini; overskirt \\
         \bottomrule
    \end{tabular}
    \caption{Mapping between 9 ambiguous ON classes and the possible correspondences in IN. Different IN classes are separated with a semicolon.
    }
    \label{ON:ambiguous}
\end{table}

\subsection{Overview of models in torchvision}

In Table~\ref{tab:model_count}, we provide a list of the models we evaluate in the main paper, along with numbers of trainable parameters and BN parameters. Note that the fraction of BN parameters is at most at 1\% compared to all trainable parameters in all considered models.

\begin{table}[htp!]
\small
\begin{center}
\begin{tabular}{lrrr}%
\toprule
Model & Parameter Count & BN Parameters & Fraction (\%)\\
\midrule
\begin{filecontents*}{data/parameters.csv}
model,params,bnparams,fraction
densenet121,7978856,83648,0.010483708441410649
densenet161,28681000,219936,0.00766835187057634
densenet169,14149480,158400,0.011194757687208293
densenet201,20013928,229056,0.011444829820512994
googlenet,13004888,15072,0.0011589488506167835
inception-v3,27161264,36224,0.0013336640003204564
mnasnet0-5,2218512,20624,0.009296321137771624
mnasnet0-75,3170208,29808,0.009402537625291463
mnasnet1-0,4383312,37920,0.008650992674032786
mnasnet1-3,6282256,48816,0.007770456982332461
mobilenet-v2,3504872,34112,0.009732737743346976
resnet101,44549160,105344,0.0023646686043014052
resnet152,60192808,151424,0.002515649377912391
resnet18,11689512,9600,0.0008212489965363824
resnet34,21797672,17024,0.0007810008334835023
resnet50,25557032,53120,0.002078488613231771
resnext101-32x8d,88791336,202880,0.002284907617563047
25028904,68224,0.0027258085292108675
shufflenet-v2-x0-5,1366792,7952,0.005818003031917073
shufflenet-v2-x1-0,2278604,16180,0.007100838934716168
shufflenet-v2-x1-5,3503624,23440,0.006690215616744263
shufflenet-v2-x2-0,7393996,33716,0.004559915910151966
vgg11-bn,132868840,5504,4.1424309868288155e-05
vgg13-bn,133053736,5888,4.4252797230736914e-05
vgg16-bn,138365992,8448,6.105546513192346e-05
vgg19-bn,143678248,11008,7.661563356479681e-05
wide-resnet101-2,126886696,137856,0.00108644959909745
wide-resnet50-2,68883240,68224,0.0009904296023241649
\end{filecontents*}
\csvreader[head to column names]{data/parameters.csv}{}
{\model 
& \num[round-mode=places,
       round-precision=2,
       scientific-notation=true]{\params} 
& \num[round-mode=places,
       round-precision=2,
       scientific-notation=true]{\bnparams} 
& \num[round-mode=places,
round-precision=3]{\fraction} \\}\\
\bottomrule
\end{tabular}
\end{center}
\caption{%
Overview of different models with parameter counts.
We show the total number of BN parameters,
which is a sum of affine parameters.
}
\label{tab:model_count}
\end{table}

\subsection{Baseline corruption errors}\label{sec:appendix-mce}

In Table~\ref{tab:alexnet}, we report the scores used for converting top-1 error into the mean corruption error (mCE) metric proposed by \citet{hendrycks2018benchmarking}.

\begin{table}[hp!]
    \small
    \centering
    \begin{tabular}{l l c }
        \toprule
        Category & Corruption & top1 error \\
        \midrule
        \multirow{3}{*}{Noise} & Gaussian Noise & 0.886428 \\
        & Shot Noise & 0.894468 \\
        & Impulse Noise & 0.922640 \\
        \midrule
        \multirow{4}{*}{Blur}& Defocus Blur & 0.819880 \\
        & Glass Blur & 0.826268 \\
        & Motion Blur & 0.785948 \\
        & Zoom Blur & 0.798360 \\
        \midrule
        \multirow{4}{*}{Weather}& Snow & 0.866816 \\
        & Frost & 0.826572 \\
        & Fog & 0.819324 \\
        & Brightness & 0.564592 \\
        & Contrast & 0.853204 \\
        \midrule
        \multirow{4}{*}{Digital}&Elastic Transform & 0.646056 \\
        &Pixelate & 0.717840 \\
        &JPEG Compression & 0.606500 \\
        \midrule
        Hold-out Noise & Speckle Noise & 0.845388 \\
        Hold-out Digital & Saturate & 0.658248 \\
        Hold-out Blur & Gaussian Blur & 0.787108 \\
        Hold-out Weather & Spatter & 0.717512 \\
        \bottomrule
    \end{tabular}
    \caption{AlexNet top1 errors on ImageNet-C
    }
    \label{tab:alexnet}
\end{table}

\subsection{Software stack}

We use various open source software packages for our experiments, most notably Docker \citep{10.5555/2600239.2600241},
scipy and numpy \citep{2020SciPy-NMeth},
GNU parallel \citep{Tange2011a},
Tensorflow \citep{abadi2016tensorflow},
PyTorch \citep{paszke2017automatic} and
torchvision \citep{10.1145/1873951.1874254}.

\clearpage
\section{Additional results}
\label{app:results}

\subsection{Performance of SimCLRv2 models}
We evaluate the performance of 3 models from the SimCLRv2 framework with and without batchnorm adaptation. We test a ResNet50, a ResNet101 and a ResNet152, finetuned on 100\% of IN training data. Since our code-base is in PyTorch, we use the Pytorch-SimCLR-Converter \citep{convert-pytorch-simclr} to convert the provided checkpoints from Tensorflow to PyTorch. We notice a slight decline in performance when comparing the top-1 accuracy on the IN validation set, see Table~\ref{app:sim-clr-v2-IN}. For preprocessing, we disable the usual PyTorch normalization and use the PIL.Image.BICUBIC interpolation for resizing because this interpolation is used in the TensorFlow code (instead of the default PIL.Image.BILINEAR in PyTorch).

The BN adaptation results for the converted models are shown in Table~\ref{app:sim-clr-v2}. Adaptation improves the performance of the ResNet50 and the ResNet101 model, but hurts the performance of the ResNet152 model. 

\newcommand{\insrulesSimCLR}{%
        \cmidrule( r){1-1}
        \cmidrule(lr){2-4}
}

\newcommand{\insrulesSimCLRIN}{%
        \cmidrule( r){1-1}
        \cmidrule(lr){2-3}
}

\begin{table}[tpb]
\begin{minipage}{.5\linewidth}

      \centering
    \small
    \captionsetup{width=.75\textwidth}
    \caption{%
After converting the checkpoints from TensorFlow to Pytorch, we notice a slight degradation in performance on the IN val set.
    }
\begin{tabular}[t]{lrr}
\multicolumn{3}{l}{%
IN val top-1 accuracy in \%.
}
\\
\toprule
Model & TF & PyTorch  \\
\insrulesSimCLRIN
SimCLRv2 ResNet50   &   76.3    &  75.6  \\
SimCLRv2 ResNet101   &   78.2    &  77.5  \\
SimCLRv2 ResNet152   &   79.3     &  78.6 \\
\bottomrule
\end{tabular}
    \label{app:sim-clr-v2-IN}

    \end{minipage}%
    \begin{minipage}{.5\linewidth}
      \centering
    \small
    \captionsetup{width=.75\textwidth}
    \caption{%
Adaptation improves the performance of the ResNet50 and the ResNet101 model but hurts the performance of the ResNet152 model.
    }
\begin{tabular}[t]{lrcr}
\multicolumn{3}{l}{%
ImageNet-C (n=4096), mCE.
}
\\
\toprule
Model, adaptation: & base & adapt & $\Delta$ \\
\insrulesSimCLR
SimCLRv2 ResNet50   &   72.4    &  68.0 &   -4.2 \\
SimCLRv2 ResNet101   &   66.6     &  65.1 &   -0.9 \\
SimCLRv2 ResNet152  &   63.7     &  64.2 &   +0.5 \\
\bottomrule
\end{tabular}
    \label{app:sim-clr-v2}
   \end{minipage}%
\end{table}

\subsection{Relationship between parameter count and IN-C improvements}

In addition to Fig.~3 in the main paper, we show the relationship between parameter count and IN-C mCE. In general, we see that the parameter counts correlates with corruption robustness since larger models have smaller mCE values.

\begin{figure}[tp]
\begin{center}
\begin{tikzpicture}
\begin{axis}[pairedscatter,
    width=0.85\textwidth,
    height=0.45\textwidth,
    xlabel={Parameter Count},
    ylabel={IN-C mCE},
    xmin=1300000,
    xmax=150000000,
    xmode=log,
    legend style={draw = none, at={(1.25,1), anchor=west}, name=legend, font=\tiny}
]

\foreach \model in {densenet,
googlenet,inception,mnasnet,mobilenet,resnet,resnext,shufflenet,vgg,wide} {

\addplot+ [%
	mark size = 0.15em,
	scatter/classes={%
        True={mark=*},
        False={mark=o}%
    },
	scatter,
	scatter src=explicit symbolic,
	discard if not={modelclass}{\model}%
] table [
    x=params,
    y=inc,
    meta=trainmode
]{data/imagenet-c/torchvision-reordered.tsv};

}
\legend{%
{ }, DenseNet,
{ }, GoogLeNet,
{ }, Inception,
{ }, MNASnet,
{ }, Mobilenet,
{ }, ResNet,
{ }, ResNext,
{ }, ShuffleNet,
{ }, VGG,
{ }, WRN,
}
\end{axis}
\end{tikzpicture}
\end{center}
\caption{%
Adaptation ($\bullet$) improves baseline ($\circ$) mCE across all 25 model architectures in the \texttt{torchvision} library, often on the order of 10\% points.
Best viewed in color.
}
\label{fig:torchvision-params}
\end{figure}
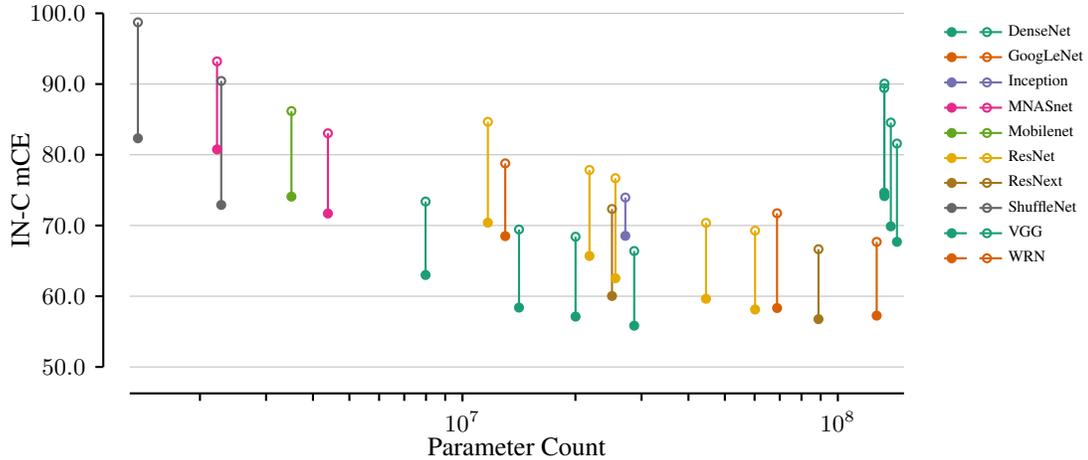

\subsection{Per-corruption results on IN-C}

We provide more detailed results on the individual corruptions of IN-C for the most important models considered in our study in Fig.~\ref{fig:resnet_augmix_individual}.
The results are shown for models where the BN parameters are adapted on the full test sets.
The adaptation consistently improves the error rates on all corruptions for both vanilla and AugMix.

\pgfkeys{/pgfplots/corruptionchart/.style={
    ybar,
    tuftelike,
    width=0.95\textwidth,
    height=0.25\textwidth,
    bar width=0.1cm,
    cycle list/Paired,
    ylabel={top-1 error},
    axis x line shift = 1pt,
    xtick=data,
    ymin=0,
    ymajorgrids, 
    major grid style={draw=white},
    ymin =  0,
    ymax = 80,
    xticklabel style={
      rotate=45,font=\tiny
    },
    symbolic x coords = {
      contrast,
      brightness,
      defocus-blur,
      elastic-transform,
      fog,
      frost,
      gaussian-blur,
      gaussian-noise,
      glass-blur,
      impulse-noise,
      jpeg-compression,
      motion-blur,
      pixelate,
      saturate,
      shot-noise,
      snow,
      spatter,
      speckle-noise,
      zoom-blur%
    },
    nodes near coords={
        \pgfmathprintnumber[precision=0]{\pgfplotspointmeta}
    },
    nodes near coords align={vertical},
    nodes near coords style={font=\tiny\sffamily, color=black},
    legend style={draw = none, at={(1.1,0.99)}, name=legend, font=\tiny}
}}

\begin{figure}[hp!]
\centering
\begin{tikzpicture}
\begin{axis}[
    corruptionchart,
    title={Vanilla Resnet-50}
    ]

\addplot+ [
  fill,
	discard if not={mode}{baseline},
] table[x = corruption, y = error] {data/imagenet-c/corruptions-rn50.tsv};

\addplot+ [
  fill,
	discard if not={mode}{adapted},
] table[x = corruption, y = error] {data/imagenet-c/corruptions-rn50.tsv};

\legend{Baseline, Adapted};
\end{axis}
\end{tikzpicture}

\begin{tikzpicture}
\begin{axis}[
    corruptionchart,
    at={(0,0.3\textwidth)},
    title={Augmix}
]

\addplot+ [
  fill,
	discard if not={mode}{baseline},
] table[x = corruption, y = error] {data/imagenet-c/corruptions-am50.tsv};

\addplot+ [
  fill,
	discard if not={mode}{adapted},
] table[x = corruption, y = error] {data/imagenet-c/corruptions-am50.tsv};

\legend{Baseline, Adapted};
\end{axis}
\end{tikzpicture}

\caption{Results on the individual corruptions of IN-C for the vanilla trained ResNet-50 and the AugMix model with and without adaptation. Adaptation reduces the error on all corruptions.} \label{fig:resnet_augmix_individual}

\end{figure}
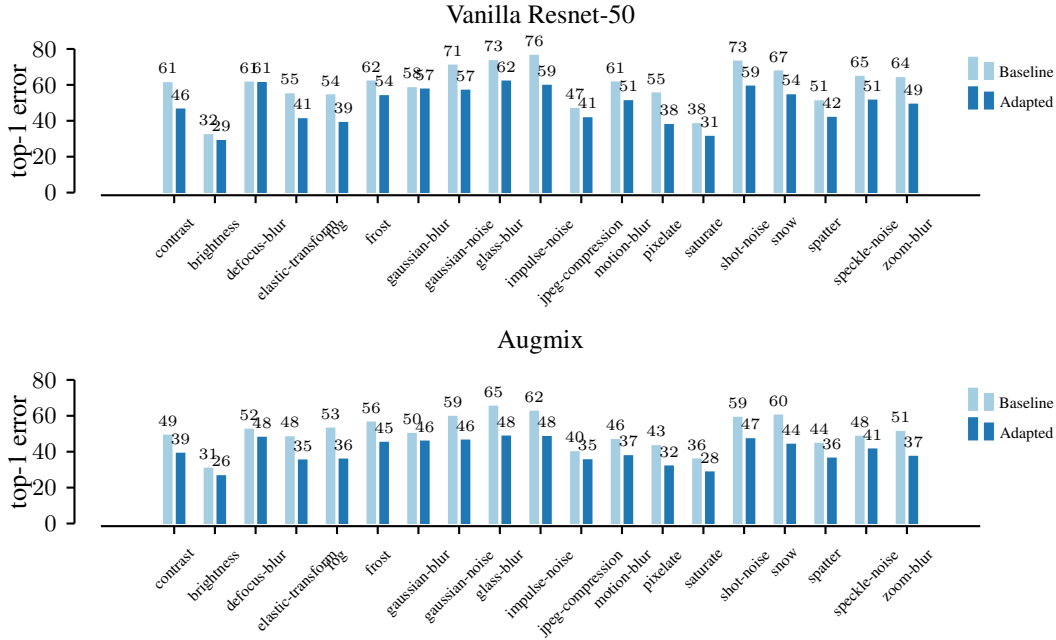

\subsection{Qualitative analysis of similarities between common corruptions}

In this analysis, we compute a t-SNE embedding of the Wasserstein distances between the adapted models and the non-adapted model from Section 5, Fig. 4(i) of the main paper.
The results are displayed in Fig.~\ref{fig:wasserstein-tnse}.
We observe that the different corruption categories indicated by the different colors are grouped together except for the 'digital' category (pink).
This visualization shows that corruption categories mostly induce similar shifts in the BN parameters. This might be an explanation why training a model on Gaussian noise generalizes so well to other noise types as has been observed by \citet{Rusak2020IncreasingTR}: By training on Gaussian noise, the BN statistics are adapted to the Gaussian noise corruption and from Fig.~\ref{fig:wasserstein-tnse}, we observe that these statistics are similar to the BN statistics of other noises.

\pgfplotsset{cycle list/Dark2}

\begin{figure}[htp!]
\centering
\begin{tikzpicture}
\begin{axis}[%
    tuftelike,
    width=0.4\textwidth,
    height=0.4\textwidth,
    title={t-SNE embedding},
    cycle multi list={%
        Dark2\nextlist
        mark=star,mark=triangle,mark=square,mark=o
    },
    legend columns=2,
    legend style={draw = none, at={(1.8,1), anchor=west}, name=legend, font=\tiny}
]

\foreach \corruption in {%
    gaussian-noise, shot-noise, impulse-noise,speckle-noise,
    defocus-blur, glass-blur, motion-blur, zoom-blur,
    snow, frost, fog, brightness,
    contrast, elastic-transform, pixelate, jpeg-compression%
    }{
\addplot+[
    discard if not={corruption}{\corruption}%
    ] table [
    x=x, 
    y=y] {data/analysis/analysis-wasserstein-tsne.tsv};

}
\legend{%
    gaussian-noise, shot-noise, impulse-noise,speckle-noise,
    defocus-blur, glass-blur, motion-blur, zoom-blur,
    snow, frost, fog, brightness,
    contrast, elastic-transform, pixelate, jpeg-compression,
    gaussian-blur, spatter, saturate
}
\end{axis}
\end{tikzpicture}
\caption{t-SNE embeddings of the Wasserstein distances between BN statistics adapted on the different corruptions. This plot shows evidence on the similarities between different corruption types.}
\label{fig:wasserstein-tnse}
\end{figure}
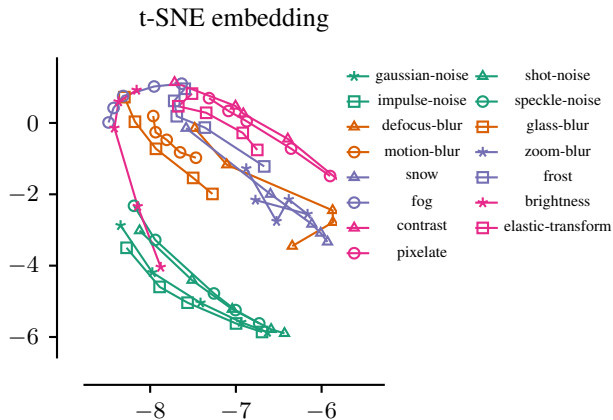

\subsection{Error prediction based on the Wasserstein distance}
\label{app:error_prediction}

In Section 5, Fig.~4(i), we observe that the relationship between the Wasserstein distance and the top-1 error on IN-C is strikingly linear in the considered range of the Wasserstein distance.
Similar corruptions and corruption types (indicated by color) exhibit similar slope, allowing to approximate the expected top-1 error rate without any information about the test domain itself.
Using the split of the 19 corruptions into 15 test and 4 holdout corruptions \citep{hendrycks2018benchmarking}, we compute a linear regression model on the five data points we get for each of the holdout corruptions (corresponding to the five severity levels), and use this model to predict the expected top-1 error rates for the remaining corruptions within the corruption family. This scheme works particularly for the ``well defined'' corruption types such as noise and digital (\num{4.14}\% points absolute mean deviation from the real error. The full results are depicted in Table~\ref{tbl:predict-error}.

\begin{table}[hp!]
\begin{center}
\small
\begin{tabular}{lrrrrrrrr}
\toprule
{} & \multicolumn{3}{c}{test error} & \multicolumn{3}{c}{holdout (train) error} & \multicolumn{2}{c}{model}\\
{} &  true &  pred  &  $|\Delta|$  &  true &  pred &  $|\Delta|$ &   coef &  intercept \\
\midrule
Fig.~\ref{fig:wasserstein-analysis-detailed} (i) \\
blur    &         64.89 &        54.53 &        11.04 &        58.13 &         58.13 &          3.24 &  37.59 &      -0.70 \\
digital &         54.37 &        51.96 &         6.97 &        38.08 &         38.08 &          0.60 &  37.20 &       6.39 \\
noise   &         73.29 &        69.68 &         5.84 &        64.51 &         64.51 &          0.65 &  24.66 &       1.68 \\
weather &         53.87 &        42.92 &        11.21 &        50.84 &         50.84 &          5.48 &  25.80 &       6.33 \\
\midrule
Fig.~\ref{fig:wasserstein-analysis-detailed} (ii) \\
blur    &         55.68 &        53.28 &         5.65 &        57.38 &         57.38 &          4.01 &  42.74 &      -9.51 \\
digital &         41.53 &        39.80 &         4.14 &        31.05 &         31.05 &          0.34 &  23.44 &      11.09 \\
noise   &         58.43 &        55.04 &         4.14 &        51.24 &         51.24 &          1.01 &  18.13 &       5.06 \\
weather &         43.84 &        36.16 &         7.80 &        41.63 &         41.63 &          4.32 &  17.80 &      10.91 \\
\midrule
Fig.~\ref{fig:wasserstein-analysis-detailed} (iii) \\
blur    &         57.10 &        69.84 &        13.43 &        74.01 &         74.01 &          3.96 &  43.50 &       5.93 \\
digital &         46.16 &        38.06 &        12.97 &        36.22 &         36.22 &         10.52 &   4.94 &      32.01 \\
noise   &         93.60 &        85.84 &        13.08 &        81.10 &         81.10 &          3.52 &  22.56 &      23.65 \\
weather &         43.74 &        36.90 &         8.98 &        44.05 &         44.05 &          6.20 &  23.29 &       3.87 \\
\bottomrule
\end{tabular}
\caption{
    Estimating top-1 error of unseen corruptions within the different corruption classes.
    We note that especially for well defined corruptions (like noise or digital corruptions), the estimation scheme works well.
    We follow the categorization originally proposed by \citet{hendrycks2018benchmarking}.
}
\label{tbl:predict-error}
\end{center}
\end{table}

\subsection{Training details on the models trained with Fixup initialization and GroupNorm}
\label{app:fixup}

In Section 5 of the main paper, we consider IN models trained with GroupNorm and Fixup initialization.
For these models, we consider the original reference implementations provided by the authors.
We train ResNet-50, ResNet-101 and ResNet-152 models with stochastic gradient descent with momentum (learning rate 0.1, momentum 0.9), with batch size 256 and weight decay \num{1e-4} for 100 epochs.

\pgfplotsset{every tick label/.append style={font=\footnotesize}}
\pgfmathsetlengthmacro\MajorTickLength{
  \pgfkeysvalueof{/pgfplots/major tick length} * 4
}

\pgfkeys{/pgfplots/priorplot/.style={
    width=0.4\textwidth,
    height=0.27\textwidth,
    cycle list/RdPu-9,
    xlabel=Batch size,
    ylabel={mCE},
    xtick=      {1,2,4,8,16,32,64,128,256,512},
    xticklabels={1, , ,8,  ,  ,64,   ,   ,512},
    xmin=1,
    xmax=512,
    ymin=50,
    ymax=100,
    ytick = {50, 60, 70, 80, 90, 100},
    legend pos=north east,
    legend style={draw = none, at={(1.5,1.3), font=\footnotesize}, name=legend},
    legend image post style={line width = 3pt}
}}

\pgfplotsset{cycle list/Dark2-8}

\subsection{Effect of Pseudo Batchsize}
\begin{figure}[hp]
\begin{center}
\begin{tikzpicture}
\begin{axis}[%
    tuftelike,
    priorplot,
    xmode=log,
    xmin=1,
    xmax=512,
    ymin=50,
    ymax=75,
    legend pos=north east,
    legend columns=3,
    legend style={draw = none, at={(1.2,0.99), font=\tiny}, name=legend},
    cycle list/Dark2-8,
   title={Performance for optimal $N$}]

    \foreach \modelname in {resnet, augmix, ant, antsin, sin}
    {
        \addplot+ [%
            opacity=.95,
            mark options = {solid},
            discard if not={model}{\modelname}%
        ] table [
            x=bsz,
            y=mce,
        ]{data/batchsize/priors.tsv};
    }
    \legend{
        ResNet,
        AugMix,
        ANT,
        ANT+SIN,
        SIN
    }
\end{axis}

\begin{axis}[%
    tuftelike,
    at={(.5\textwidth,0)},
    priorplot,
    xmin=1,
    xmax=512,
    ymin=8,
    ymax=1024,
    ylabel={Pseudo Batch Size},
    ymode=log,
    xmode=log,
    ytick=      {1,2,4,8,16,32,64,128,256,512},
    yticklabels={1, , ,8,  ,  ,64,   ,   ,512},
    legend pos=north east,
    legend columns=3,
    legend style={draw = none, at={(1.1,0.99), font=\footnotesize}, name=legend},
    cycle list/Dark2-8,
   title={Best Pseudo Batchsize $N$}]
    \foreach \modelname/\idx in {resnet/0, augmix/1, ant/2, antsin/3, sin/4}
    {
        \addplot+ [%
            opacity=.95,
            mark options = {solid},
        ] table [
            x=bsz,
            y=\modelname,
        ]{data/batchsize/priors-optimal-N.tsv};
    }
\end{axis}
\end{tikzpicture}
\end{center}
\caption{%
Left: Performance for all the considered ResNet-50 variants based on the sample batch size.
The optimal $N$ is chosen according to the mCE on the holdout corruptions.
Right: Best choice for $N$ depending on the input batchsize $n$.
Note that in general for high values $n$, the model is generally more robust to the choice of $N$.
}

\begin{center}
\begin{tikzpicture}

\foreach \modelname/\ddx/\ddy in {resnet/0/0} {
    \begin{axis}[tuftelike,priorplot,xmode=log,title=\modelname,at={(\ddx*0.5*\textwidth,-\ddy*0.31*\textwidth)}]
    \addlegendimage{empty legend}
    \foreach \ccol/\oppp in {
        1/20,
        2/30,
        4/40,
        8/50,
        16/60,
        32/70,
        64/80,
        128/90,
        256/100
    }
    {
        \edef\temp{\noexpand\addplot+ [
            black!\oppp,
            mark=none
            ] table [x=batchsize, y=\ccol]{data/priors/\modelname.tsv};
        }
        \temp
    }
    \legend{
        \hspace{-.6cm}$N$, 
        1, 2, 4, 8, 16, 32, 64, 128, 256, 512
    }
    \end{axis}
}

\foreach \modelname/\ddx/\ddy in {augmix/1/1,
                        sin/0/1, ant/1/2, antsin/0/2} {
    \begin{axis}[tuftelike,priorplot,xmode=log,title=\modelname,at={(\ddx*0.5*\textwidth,-\ddy*0.31*\textwidth)}]
    \addlegendimage{empty legend}
    \foreach \ccol/\oppp in {
        1/20,
        2/30,
        4/40,
        8/50,
        16/60,
        32/70,
        64/80,
        128/90,
        256/100
    }
    {
        \edef\temp{\noexpand\addplot+ [
            black,
            black!\oppp,
            mark=none
            ] table [x=batchsize, y=\ccol]{data/priors/\modelname.tsv};
        }
        \temp
    }
    \end{axis}
}

\end{tikzpicture}
\end{center}
\caption{%
Effects of batch size $n$ and pseudo batch size $N$ for the various considered models.
We report mCE averaged across 15 test corruptions.
}
\label{fig:bsz_ant}
\end{figure}
We show the full results for considering different choices of $N$ for ResNet-50, Augmix, ANT, ANT+SIN and SIN models and display the result in Fig.~\ref{fig:bsz_ant}.
We observe a characteristic shape which we believe can be attributed to the way statistics are estimated.
We provide evidence for this view by proposing an analytical model which we discuss in \textsection\ref{app:analytical}.

\begin{table}
\scriptsize
\centering
\begin{tabular}{lrrrrrrrrr}
\toprule
ResNet-50 &     1   &    2   &    4   &    8   &    16  &    32  &    64  &    128 &    256 \\
\midrule
1         &  117.76 &  98.78 &  81.06 &  72.80 &  71.39 &  72.72 &  74.28 &  75.36 &  75.99 \\
2         &   98.11 &  89.92 &  80.13 &  72.36 &  69.63 &  70.39 &  72.39 &  74.16 &  75.32 \\
4         &   81.10 &  78.45 &  74.70 &  70.27 &  67.48 &  67.69 &  69.77 &  72.19 &  74.10 \\
8         &   71.56 &  70.74 &  69.44 &  67.56 &  65.60 &  65.02 &  66.70 &  69.41 &  72.07 \\
16        &   66.82 &  66.52 &  66.06 &  65.32 &  64.29 &  63.32 &  63.81 &  66.19 &  69.24 \\
32        &   64.51 &  64.39 &  64.19 &  63.87 &  63.38 &  62.72 &  62.21 &  63.22 &  65.94 \\
64        &   63.33 &  63.28 &  63.19 &  63.05 &  62.81 &  62.43 &  61.95 &  61.68 &  62.90 \\
128       &   62.78 &  62.75 &  62.69 &  62.62 &  62.50 &  62.29 &  62.00 &  61.56 &  61.42 \\
256       &   62.51 &  62.49 &  62.44 &  62.41 &  62.32 &  62.22 &  62.01 &  61.73 &  61.35 \\
512       &   62.36 &  62.36 &  62.33 &  62.29 &  62.26 &  62.17 &  62.06 &  61.90 &  61.62 \\
\midrule
\midrule
AugMix &     1   &    2   &    4   &    8   &    16  &    32  &    64  &    128 &    256 \\
\midrule
1         &  122.56 &  99.72 &  76.23 &  65.46 &  62.08 &  61.78 &  62.70 &  63.75 &  64.47 \\
2         &  100.39 &  88.69 &  75.16 &  64.86 &  60.93 &  60.51 &  61.28 &  62.52 &  63.67 \\
4         &   78.55 &  74.41 &  68.69 &  62.52 &  58.58 &  58.30 &  59.53 &  60.94 &  62.39 \\
8         &   65.02 &  63.81 &  61.86 &  59.21 &  56.39 &  55.40 &  56.87 &  59.00 &  60.77 \\
16        &   58.02 &  57.55 &  56.96 &  56.02 &  54.69 &  53.44 &  53.78 &  56.15 &  58.71 \\
32        &   54.37 &  54.20 &  53.99 &  53.68 &  53.21 &  52.50 &  51.99 &  53.01 &  55.78 \\
64        &   52.55 &  52.50 &  52.38 &  52.24 &  52.07 &  51.83 &  51.39 &  51.25 &  52.59 \\
128       &   51.64 &  51.60 &  51.54 &  51.47 &  51.38 &  51.26 &  51.10 &  50.88 &  50.89 \\
256       &   51.18 &  51.17 &  51.12 &  51.08 &  51.02 &  50.95 &  50.86 &  50.76 &  50.60 \\
512       &   50.96 &  50.95 &  50.93 &  50.90 &  50.86 &  50.80 &  50.72 &  50.65 &  50.61 \\
\midrule
\midrule
ANT &     1   &    2   &    4   &    8   &    16  &    32  &    64  &    128 &    256 \\
\midrule
1         &  116.10 &  93.58 &  72.31 &  62.28 &  60.07 &  60.73 &  61.75 &  62.48 &  62.90 \\
2         &   93.88 &  83.74 &  72.01 &  62.69 &  58.97 &  59.10 &  60.44 &  61.67 &  62.44 \\
4         &   74.51 &  71.06 &  66.34 &  61.15 &  57.55 &  57.03 &  58.51 &  60.29 &  61.64 \\
8         &   63.65 &  62.50 &  60.74 &  58.43 &  56.04 &  55.02 &  56.10 &  58.22 &  60.20 \\
16        &   58.37 &  57.87 &  57.14 &  56.11 &  54.77 &  53.67 &  53.76 &  55.61 &  58.06 \\
32        &   55.78 &  55.54 &  55.20 &  54.66 &  53.91 &  53.06 &  52.50 &  53.18 &  55.35 \\
64        &   54.51 &  54.41 &  54.21 &  53.88 &  53.42 &  52.84 &  52.23 &  51.94 &  52.87 \\
128       &   53.92 &  53.85 &  53.71 &  53.53 &  53.28 &  52.85 &  52.29 &  51.80 &  51.65 \\
256       &   53.66 &  53.61 &  53.50 &  53.37 &  53.20 &  52.96 &  52.54 &  52.04 &  51.60 \\
512       &   53.53 &  53.49 &  53.41 &  53.33 &  53.21 &  53.02 &  52.78 &  52.38 &  51.90 \\
\midrule
\midrule
ANT+SIN &     1   &    2   &    4   &    8   &    16  &    32  &    64  &    128 &    256 \\
\midrule
1         &  108.24 &  84.75 &  67.42 &  59.91 &  58.15 &  58.49 &  59.24 &  59.85 &  60.23 \\
2         &   87.60 &  78.40 &  68.32 &  60.63 &  57.54 &  57.47 &  58.33 &  59.23 &  59.87 \\
4         &   71.12 &  68.32 &  64.31 &  59.78 &  56.63 &  56.06 &  57.01 &  58.24 &  59.23 \\
8         &   62.23 &  61.38 &  59.98 &  57.93 &  55.69 &  54.59 &  55.30 &  56.79 &  58.21 \\
16        &   57.83 &  57.51 &  57.00 &  56.17 &  54.96 &  53.76 &  53.61 &  54.92 &  56.68 \\
32        &   55.62 &  55.51 &  55.33 &  54.96 &  54.38 &  53.55 &  52.80 &  53.13 &  54.73 \\
64        &   54.57 &  54.49 &  54.40 &  54.25 &  53.98 &  53.51 &  52.84 &  52.36 &  52.89 \\
128       &   54.02 &  53.98 &  53.95 &  53.85 &  53.72 &  53.49 &  53.07 &  52.53 &  52.12 \\
256       &   53.76 &  53.74 &  53.71 &  53.67 &  53.59 &  53.47 &  53.23 &  52.85 &  52.33 \\
512       &   53.64 &  53.63 &  53.60 &  53.57 &  53.51 &  53.45 &  53.35 &  53.12 &  52.75 \\
\midrule
\midrule
SIN &     1   &    2   &    4   &    8   &    16  &    32  &    64  &    128 &    256 \\
\midrule
1         &  119.11 &  94.43 &  74.93 &  67.03 &  65.43 &  66.08 &  67.16 &  68.04 &  68.62 \\
2         &   98.85 &  88.62 &  76.99 &  67.88 &  64.23 &  64.42 &  65.72 &  67.02 &  67.99 \\
4         &   81.35 &  78.10 &  73.38 &  67.84 &  63.49 &  62.47 &  63.76 &  65.48 &  66.94 \\
8         &   70.92 &  69.94 &  68.38 &  66.02 &  63.14 &  61.09 &  61.45 &  63.35 &  65.35 \\
16        &   65.29 &  64.97 &  64.48 &  63.68 &  62.39 &  60.78 &  59.90 &  60.92 &  63.16 \\
32        &   62.34 &  62.25 &  62.08 &  61.80 &  61.36 &  60.55 &  59.55 &  59.26 &  60.65 \\
64        &   60.84 &  60.80 &  60.74 &  60.61 &  60.47 &  60.15 &  59.67 &  58.96 &  58.93 \\
128       &   60.07 &  60.04 &  60.02 &  59.96 &  59.87 &  59.77 &  59.57 &  59.18 &  58.64 \\
256       &   59.68 &  59.66 &  59.64 &  59.62 &  59.59 &  59.53 &  59.43 &  59.27 &  58.97 \\
512       &   59.48 &  59.47 &  59.46 &  59.44 &  59.42 &  59.40 &  59.33 &  59.26 &  59.11 \\
\midrule
\midrule
DeepAugment &    1   &    2   &    4   &    8   &    16  &    32  &    64  &    128 &    256 \\
\midrule
8         &  65.37 &  63.87 &  61.37 &  58.11 &  54.48 &  52.17 &  52.33 &  54.18 &  56.36 \\
\midrule
\midrule
DeepAugment+AugMix &    1   &    2   &    4   &    8   &   16  &    32  &    64  &    128 &    256 \\
\midrule
8         &  52.59 &  51.98 &  51.05 &  49.83 &  48.5 &  47.81 &  48.36 &  49.72 &  51.12 \\
\midrule
\midrule
ResNext+DeepAugment+Augmix &    1   &    2   &    4   &    8   &    16  &    32  &    64  &    128 &    256 \\
\midrule
8         &  42.09 &  41.74 &  41.29 &  40.67 &  39.96 &  39.69 &  40.35 &  41.55 &  42.69 \\
\bottomrule
\end{tabular}
\caption{Test mCE for various batch sizes (rows) vs. pseudo batch sizes (columns)}
\end{table}

\clearpage
\section{Analytical error model}\label{app:analytical}

We consider a univariate model in \textsection\ref{app:proof-sketch}--\ref{app:proof-proof} and discuss a simple extension to the multivariate diagonal case in \textsection\ref{app:proof-multi}.
As highlighted in the main text, the model qualitatively explains the overall characteristics of our experimental data.
Note that we assume a linear relationship between the Wasserstein distance and the error under domain shift, as suggested by our empirical findings.

\paragraph{Univariate model.}

We denote the source statistics as $\mu_s,\sigma_s^2$, the true target statistics as $\mu_t,\sigma^2_t$ and the 
estimated target statistics as $\hat{\mu}_t,\hat{\sigma}_t^2$.
For normalization, we take a convex combination of the source statistics and estimated target statistics:
\begin{align}\label{definitionbarmu}
    \Bar{\mu} = \frac{N}{N+n} \mu_s + \frac{n}{N+n} \hat{\mu}_t, \;
    \Bar{\sigma}^2 = \frac{N}{N+n} \sigma^2_s + \frac{n}{N+n} \hat{\sigma}^2_t .
\end{align}
We now analyze the trade-off between using an estimate closer to the source or closer to the estimated target statistics.
In the former case, the model will suffer under the covariate shift present between target and source distribution.
In the latter case, small batch sizes $n$ will yield unreliable estimates for the true target statistics, which might hurt the performance even more than the source-target mismatch.
Hence, we aim to gain understanding in the trade-off between both options, and potential optimal choices of $N$ for a given sample size $n$.

As a metric of domain shift with good properties for our following derivation, we leverage the Wasserstein distance.
In \textsection\ref{sec:results} and \textsection\ref{app:error_prediction}, we already established an empirical link between domain shift measured in terms of the top-1 performance vs. the Wasserstein distance between model statistics and observed a linear relationship for case of common corruptions.

\renewcommand\theprop{1}
\begin{prop}[Bounds on the expected value of the Wasserstein distance between target and combined estimated target and source statistics]\label{prop:bounds}

We denote the source statistics as $\mu_s,\sigma_s^2$, the true target statistics as $\mu_t,\sigma^2_t$ and the biased estimates of the target statistics as $\hat{\mu}_t,\hat{\sigma}_t^2$. 
For normalization, we take a convex combination of the source statistics and estimated target statistics as discussed in Eq.~\ref{definitionbarmu}.
At a confidence level $1-\alpha$, the expectation value of the squared Wasserstein distance
$W_2^2(\Bar{\mu}, \Bar{\sigma}, \mu_t, \sigma_t)$ between ideal and estimated target statistics w.r.t. to the distribution of sample mean $\hat{\mu}_t$ and sample variance $\hat{\sigma}^2_t$ is bounded from above and below
with $L \leq \mathbb{E}[W^2_2] \leq U$, where
\begin{equation}\label{univariatebound}
\begin{aligned}
    L &= \left( \sigma_t - \sqrt{ \frac{N}{N+n} \sigma^2_s + \frac{n-1}{N+n} \sigma^2_t } \right)^2 + \frac{N^2}{(N+n)^2} \left(\mu_t -  \mu_s \right)^2 + \frac{n}{(N+n)^2} \sigma_t^2 \\
    U &= \; L + \sigma^5_t \frac{(n-1)}{2(N+n)^2}  \left( \frac{N}{N+n} \sigma^2_s + \frac{1}{N+n} \chi^2_{1-\alpha/2, n-1} \sigma^2_t \right)^{-3/2}
\end{aligned}
\end{equation}
The quantity $\chi^2_{1-\alpha/2, n-1}$ denotes the left tail value of a chi square distribution with $n-1$ degrees of freedom, defined as $P\left(X \leq \chi^2_{1-\alpha/2, n-1}\right) = \alpha/2 \text{ for } X \sim \chi^2_{n-1}$.
\end{prop}

\subsection{Proof sketch}
\label{app:proof-sketch}

We are interested in the expected value of the Wasserstein distance defined in \eqref{def:wasserstein} between the target statistics $\mu_t, \sigma^2_t$ and the mixed statistics $\Bar{\mu}, \Bar{\sigma}^2$ introduced above in equation \eqref{definitionbarmu}, taken with respect to the distribution of the sample moments $\hat{\mu}_t$, $\hat{\sigma}_t^2$. The expectation value itself cannot be evaluated in closed form because the Wasserstein distance contains a term proportional to $\Bar{\sigma}$ being the square root of the convex combination of target and source variance. 

In Lemma \ref{lemma1}, the square root term is bounded from above and below using Jensen's inequality and Holder's defect formula which is reviewed in Lemma \ref{holdersdefect}. After having bounded the problematic square root term, the proof of Proposition~\ref{prop:bounds} reduces to inserting the expectation values of sample mean and sample variance reviewed in Lemma \ref{lemma:moments}.

\subsection{Prerequisites}
\label{app:proof-reqisisites}

\begin{lemma}[Mean and variance of sample moments, following \citep{weisstein}]\label{lemma:moments}
The sample moments $\hat{\mu}_t,\hat{\sigma}_t^2$ are random variables depending on the sample size $n$.
\begin{align}
    \hat{\mu}_t &= \; \frac{1}{n} \sum_{j=1}^n x_j, \quad
    \hat{\sigma}_t^2 = \; \frac{1}{n} \sum_{j=1}^n (x_j - \hat{\mu}_t)^2 \; \text{ with } x_j \sim \normaldist \left(\mu_t, \sigma^2_t \right).
\end{align}
For brevity, we use the shorthand $\mathbb{E}[\cdot]$ for all expectation values with respect to the distribution of $p(\hat{\mu}_t,\hat{\sigma}_t^2|n)$.
In particular, our computation uses mean and variance of $\hat{\mu}_t$ and $\hat{\sigma}_t^2$ which are well known for a normal target distribution:
\begin{align}
    \hat{\mu}_t \sim \normaldist\left(\mu_t, \frac{1}{n} \sigma^2_t \right), \; \mathbb{E}[\hat{\mu}_t] &= \; \mu_t, \;
    \mathbb{V}[\hat{\mu}_t] = \frac{1}{n} \sigma^2_t \\
    \frac{\hat{\sigma}^2_t}{\sigma_t^2/n} \sim \chi^2_{n-1}, \; \mathbb{E}[ \hat{\sigma}_t^2] &= \; \frac{n-1}{n} \sigma^2_t, \; \mathbb{V}[\hat{\sigma}^2_t] = \frac{\sigma_t^4}{n^2} \mathbb{V}\left[\frac{\hat{\sigma}^2_t}{\sigma_t^2/n}\right] =
    \frac{\sigma_t^4}{n^2} \, 2(n-1).
\end{align}
The derivation of the variance $\mathbb{V}[\hat{\sigma}_t^2]$ in the last line uses the fact that the variance of a chi square distributed variable with $(n-1)$ degrees of freedom is equal to $2(n-1)$.
\end{lemma}

\begin{lemma}[Holder's defect formula for concave functions in probabilistic notation, following \citet{becker2012variance} ] \label{holdersdefect}
If the concave function $f: [a,b] \to \mathbb{R}$ is twice continuously differentiable and there are finite bounds $m$ and $M$ such that
\begin{align}
    -M \leq f''(x) \leq -m \leq 0 \; \forall x \in [a,b],
\end{align}
then the defect between Jensen's inequality estimate $f\left(\mathbb{E}[X]\right)$ for a random variable $X$ taking values $x \in [a,b]$ and the true expectation value $\mathbb{E}[f(X)]$ is bounded from above by a term proportional to the variance of $X$:
\begin{align}
    f \left( \mathbb{E}[X] \right) - \mathbb{E}[f(X)] \leq \frac{1}{2} M \mathbb{V}[X].
\end{align}
\end{lemma}

\begin{lemma}[Upper and lower bounds on the expectation value of $\Bar{\sigma}$] \label{lemma1}
The expectation value of the square root of the random variable $\Bar{\sigma}^2$ defined as
\begin{align}\label{definitionX}
    \Bar{\sigma}^2 &= \frac{N}{N+n} \sigma^2_s + \frac{n}{N+n} \hat{\sigma}^2_t,
\end{align}
is bounded from above and below at a confidence level $1-\alpha$ by
\begin{align}                        \sqrt{\mathbb{E}\left[ \Bar{\sigma}^2 \right     ]}  - \frac{1}{2} M               \mathbb{V}[\Bar{\sigma}^2] &\leq 
    \mathbb{E}\left[\sqrt{\Bar{\sigma}^2}\right] \leq 
     \sqrt{\mathbb{E}\left[\Bar{\sigma}^2\right]}\\
    \sqrt{\mathbb{E}\left[\Bar{\sigma}^2\right]} &= \sqrt{\frac{N}{N+n} \sigma^2_s + \frac{n-1}{N+n} \sigma^2_t}, \\
    \frac{1}{2} M \mathbb{V}[\Bar{\sigma}^2] &= \frac{(n-1)}{4(N+n)^2} \sigma^4_t \left( \frac{N}{N+n} \sigma^2_s + \frac{1}{N+n} \chi^2_{1-\alpha/2, n-1} \sigma^2_t \right).
\end{align}
The quantity $\chi^2_{1-\alpha/2, n-1}$ denotes the left tail value of a chi square distribution with $n-1$ degrees of freedom, defined as $P\left(X \leq \chi^2_{1-\alpha/2, n-1}\right) = \alpha/2 \; \text{ for } X \sim \chi^2_{n-1}$.

\begin{proof}

The square root function is concave, therefore Jensen's inequality implies the upper bound
\begin{align}
    \mathbb{E}\left[\sqrt{\Bar{\sigma}^2}\right] \leq \sqrt{\mathbb{E}[\Bar{\sigma}^2]}.
\end{align}
The square root of the expectation value of $\Bar{\sigma}^2$ is computed using the expectation value of the sample variance as given in Lemma \ref{lemma:moments}.
\begin{align}
    \sqrt{\mathbb{E}[\Bar{\sigma}^2]} &= \sqrt{\frac{N}{N+n} \sigma^2_s + \frac{n}{N+n} \frac{n-1}{n} \sigma^2_t} 
    = \sqrt{\frac{N}{N+n} \sigma^2_s + \frac{n-1}{N+n} \sigma^2_t}.
\end{align}
To state a lower bound, we use Holder's defect formula in probabilistic notation stated in Lemma \ref{holdersdefect}.
Holder's formula for concave functions requires that the random variable $\Bar{\sigma}^2$ can take values in the compact interval $[a,b]$  and that the second derivative of the square root function $f(\Bar{\sigma}^2) = \sqrt{\Bar{\sigma}^2}$, exists and is strictly smaller than zero in $[a,b]$. 
Regarding the interval of $\Bar{\sigma}^2$, we provide probabilistic upper and lower bounds. 
The ratio of sample variance and true variance divided by $n$ follows a chi  square distribution with $n-1$ degrees of freedom. 
At confidence level $1-\alpha$, this ratio lies between $\chi^2_{1-\alpha/2, n-1}$ and $\chi^2_{\alpha/2, n-1}$ which are defined as follows:
\begin{align}
    \chi^2_{1-\alpha/2, n-1} &\leq \frac{\hat{\sigma}^2_t}{\sigma^2_t/n} \leq \chi^2_{\alpha/2, n-1}, \\
    Pr(X \leq \chi^2_{1-\alpha/2, n-1}) &= \frac{\alpha}{2}, \;
    Pr(X \geq \chi^2_{\alpha/2, n-1}) = \frac{\alpha}{2}. 
\end{align}
Then at the same confidence level, the sample variance itself lies between the two quantiles multiplied by $\sigma_t^2/n$,
\begin{align}
   \chi^2_{1-\alpha/2, n-1} \frac{\sigma_t^2}{n} &\leq  \hat{\sigma}_t^2 \leq \chi^2_{\alpha/2, n-1} \frac{\sigma_t^2}{n},
\end{align}
and the random variable $\Bar{\sigma}^2$ lies in the interval
\begin{align} \label{definitionofa}
    \Bar{\sigma}^2 \in [a,b] \text{ with } a &= \frac{N}{N+n} \sigma^2_s + \frac{1}{N+n} \chi^2_{1-\alpha/2, n-1} \sigma_t^2, \\
    \text{ and } b &= \frac{N}{N+n} \sigma^2_s + \frac{1}{N+n} \chi^2_{\alpha/2, n-1} \sigma_t^2.
\end{align}
The variances and chi square values are all positive and therefore both $a$ and $b$ are positive as well, implying that the second derivative of the square root is strictly negative in the interval $[a,b]$.
\begin{align}
    f(\Bar{\sigma}^2) &= \sqrt{\Bar{\sigma}^2}, \; f'(\Bar{\sigma}^2) = \frac{1}{2} (\Bar{\sigma}^2)^{-1/2}, \; f''(\Bar{\sigma}^2) = - \frac{1}{4} (\Bar{\sigma}^2)^{-3/2} < 0 \in [a,b].
\end{align}
Consequently the second derivative is in the interval $[M, m]$ at the given confidence level:
\begin{align}\label{computeM} 
   -M &\leq f''(\Bar{\sigma}^2) \leq -m \leq 0 \text{ for } \Bar{\sigma}^2 \in [a,b] \text{ with }
   M = \frac{1}{4} a^{-3/2}, \; m = \frac{1}{4} b^{-3/2}.
\end{align}
The defect formula \ref{holdersdefect} states that the defect is bounded by
\begin{align}\label{defectformula}
    \sqrt{\mathbb{E}[\Bar{\sigma}^2]} -  \mathbb{E}[\sqrt{\Bar{\sigma}^2}]  &\leq \frac{1}{2} M \mathbb{V}[\Bar{\sigma}^2].
\end{align}
The constant $M$ was computed above in \eqref{computeM}, and the variance of $\Bar{\sigma}^2$ is calculated in the next lines, using the first and second moment of the sample variance as stated in \ref{lemma:moments}.
\begin{equation}
\begin{aligned}\label{varianceX}
    \mathbb{V}[\Bar{\sigma}^2] &= \mathbb{E}[(\Bar{\sigma}^2 - \mathbb{E}[\Bar{\sigma}^2])^2] =  \mathbb{E}\left[\left(\frac{n}{N+n} \hat{\sigma}^2_t - \frac{n}{N+n} \frac{n-1}{n}\sigma^2_t \right)^2 \right] \\
    &= \frac{n^2}{(N+n)^2} \mathbb{E}\left[ \left(\hat{\sigma}^2_t - \mathbb{E}[\hat{\sigma}^2_t\right)^2\right]
    = \frac{n^2}{(N+n)^2} \mathbb{V}\left[\hat{\sigma}^2_t \right] \\
    &= \frac{n^2}{(N+n)^2} \frac{2(n-1)}{n^2} \sigma^4_t = \frac{2(n-1)}{(N+n)^2} \sigma^4_t .
\end{aligned}
\end{equation}
Inserting $\mathbb{V}[\Bar{\sigma}^2]$ computed in \eqref{varianceX} and $M$ defined in \eqref{computeM} with $a$ as defined in \eqref{definitionofa} into the defect formula \eqref{defectformula} yields the lower bound: 
\begin{equation}
\begin{aligned}
    &\sqrt{\mathbb{E}[\Bar{\sigma}^2]} - \frac{1}{2} M \mathbb{V}[\Bar{\sigma}^2] \leq \mathbb{E}[\sqrt{\Bar{\sigma}^2}] \\
    & \sqrt{\mathbb{E}[\Bar{\sigma}^2]} - \frac{1}{2} M \mathbb{V}[\Bar{\sigma}^2] \\
    &= \sqrt{\mathbb{E}[\Bar{\sigma}^2]}  - \frac{1}{2} \cdot \frac{1}{4} a^{-3/2} \frac{2(n-1)}{(N+n)^2} \sigma^4_t \\
    &= \sqrt{\mathbb{E}[\Bar{\sigma}^2]}  -  \frac{(n-1)}{4(N+n)^2} \sigma^4_t \left(\frac{N}{N+n} \sigma^2_s + \frac{1}{N+n} \chi^2_{1-\alpha/2, n-1} \sigma_t^2\right)^{-3/2}.
\end{aligned}
\end{equation}
Assuming that source and target variance are of the same order of magnitude $\sigma$, the defect will be of order of magnitude $\sigma$: The factor $\mathbb{V}[X]$ scales with $\sigma^4$ and $M$ with $\sigma^{-3}$.
\end{proof}
\end{lemma}

\subsection{Proof of Proposition 1}
\label{app:proof-proof}

\begin{proof}
For two univariate normal distributions with moments $\mu_t, \sigma^2_t$ and $\Bar{\mu}, \Bar{\sigma}^2$, the Wasserstein distance as defined in \eqref{def:wasserstein} reduces to
\begin{align}
    W_2^2 &= \sigma^2_t + \Bar{\sigma}^2 - 2 \Bar{\sigma} \sigma_t + (\Bar{\mu} - \mu)^2.
\end{align}
The expected value of the Wasserstein distance across many batches is given as 
\begin{equation}
\begin{aligned}\label{expectedvaluewasserstein}
    \mathbb{E}[W_2^2] =& \; \sigma_t^2 +  \mathbb{E}[\Bar{\sigma}^2] - 2  \mathbb{E}[\Bar{\sigma}] \sigma_t + \mathbb{E}[(\mu_t - \Bar{\mu})^2] \\
    =& \; \sigma^2_t + \frac{N}{N+n} \sigma^2_s + \frac{n}{N+n} \frac{n-1}{n} \sigma^2_t - 2 \sigma_t \mathbb{E}\left[\sqrt{\frac{N}{N+n} \sigma^2_s + \frac{n}{N+n} \hat{\sigma}^2_t}\right] \\
    &+ \mathbb{E}\left[\left(\mu_t - \frac{N}{N+n} \mu_s - \frac{n}{N+n} \hat{\mu}_t \right)^2 \right] 
\end{aligned}
\end{equation}
which can already serve as the basis for our numerical simulations.
To arrive at a closed form analytical solution, we invoke Lemma~\ref{lemma1} to bound the expectation value $\mathbb{E}\left[\Bar{\sigma}\right]$ in equation \eqref{expectedvaluewasserstein}. 
\begin{align}
    \label{boundsquareroot}
    -2 \sigma_t \sqrt{\mathbb{E}\left[\Bar{\sigma}^2\right]} &\leq -2 \sigma_t
    \mathbb{E}\left[\sqrt{\Bar{\sigma}^2}\right] \leq 
    -2 \sigma_t \sqrt{\mathbb{E}\left[\Bar{\sigma}^2\right]} -2 \sigma_t \left( -\frac{1}{2} M \mathbb{V}[\Bar{\sigma}^2] \right)
\end{align}
Apart from the square root term bounded in equation \eqref{boundsquareroot} above, the expectation value of the Wasserstein distance can be computed exactly. Hence the bounds on $\mathbb{E}\left[\Bar{\sigma}\right]$ multiplied by a factor of $(-2 \sigma^2_t)$ coming from equation \eqref{expectedvaluewasserstein} determine lower and upper bounds $L$ and $U$ on the expected value of $W_2^2$: 
\begin{align}
     L &\leq \mathbb{E}\left[W^2_2\right] \leq U = L + \sigma_t M \mathbb{V}[\Bar{\sigma}^2]
\end{align}
In the next lines, the lower bound is calculated:
\begin{equation}
\begin{aligned} 
    L &= \; \sigma^2_t + \frac{N}{N+n} \sigma^2_s + \frac{n-1}{N+n} \sigma^2_t - 2 \sigma_t \sqrt{\mathbb{E}\left[ \frac{N}{N+n} \sigma^2_s + \frac{n-1}{N+n} \sigma^2_t \right]} \\
    &+ \left(\mu_t - \frac{N}{N+n} \mu_s\right)^2 - 2\left(\mu_t - \frac{N}{N+n} \mu_s \right) \frac{n}{N+n} \mathbb{E}[\hat{\mu}_t] + \frac{n^2}{(N+n)^2} \left( \mathbb{V}[\hat{\mu}_t] + \left(\mathbb{E}[\hat{\mu}_t] \right)^2 \right) \\ 
    = & \; \sigma^2_t +  \frac{N}{N+n} \sigma^2_s + \frac{n-1}{N+n} \sigma^2_t - 2 \sigma_t \sqrt{\frac{N}{N+n} \sigma^2_s +  \frac{n-1}{N+n} \sigma^2_t} \\
    &+ \left(\mu_t - \frac{N}{N+n} \mu_s\right)^2 - 2\left(\mu_t - \frac{N}{N+n} \mu_s \right) \frac{n}{N+n} \mu_t  +  \frac{n^2}{(N+n)^2} \left( \frac{1}{n} \sigma^2_t + \mu_t^2 \right) \\
    = & \; \left( \sigma_t - \sqrt{ \frac{N}{N+n} \sigma^2_s + \frac{n-1}{N+n} \sigma^2_t } \right)^2 + \left( \mu_t - \frac{N}{N+n} \mu_s - \frac{n}{N+n} \mu_t \right)^2 + \frac{n}{(N+n)^2} \sigma_t^2 \\
    = & \; \left( \sigma_t - \sqrt{ \frac{N}{N+n} \sigma^2_s + \frac{n-1}{N+n} \sigma^2_t } \right)^2 + \frac{N^2}{(N+n)^2} \left(\mu_t -  \mu_s \right)^2 + \frac{n}{(N+n)^2} \sigma_t^2 
\end{aligned}
\end{equation}
After having derived the lower bound, the upper bound is the sum of the lower bound and the defect term as computed in Lemma \ref{lemma1}.
\begin{equation}
    \begin{aligned}
    \mathbb{E}[W^2] &\geq U = L + \sigma_t M \mathbb{V}[\Bar{\sigma}^2] \\
    &= L + \sigma_t \frac{1}{4} \left( \frac{N}{N+n} \sigma^2_s + \frac{n}{N+n} \chi^2_{1-\alpha/2, n-1} \frac{\sigma_t^2}{n} \right)^{-3/2} \frac{2(n-1)}{(N+n)^2} \sigma_t^4 \\
    &= L + \left( \frac{N}{N+n} \sigma^2_s + \frac{1}{N+n} \chi^2_{1-\alpha/2, n-1} \sigma_t^2 \right)^{-3/2} \frac{(n-1)}{2(N+n)^2} \sigma_t^5.
\end{aligned}
\end{equation}
\end{proof}

\begin{figure}[hp]
\centering
\includegraphics[width=\textwidth]{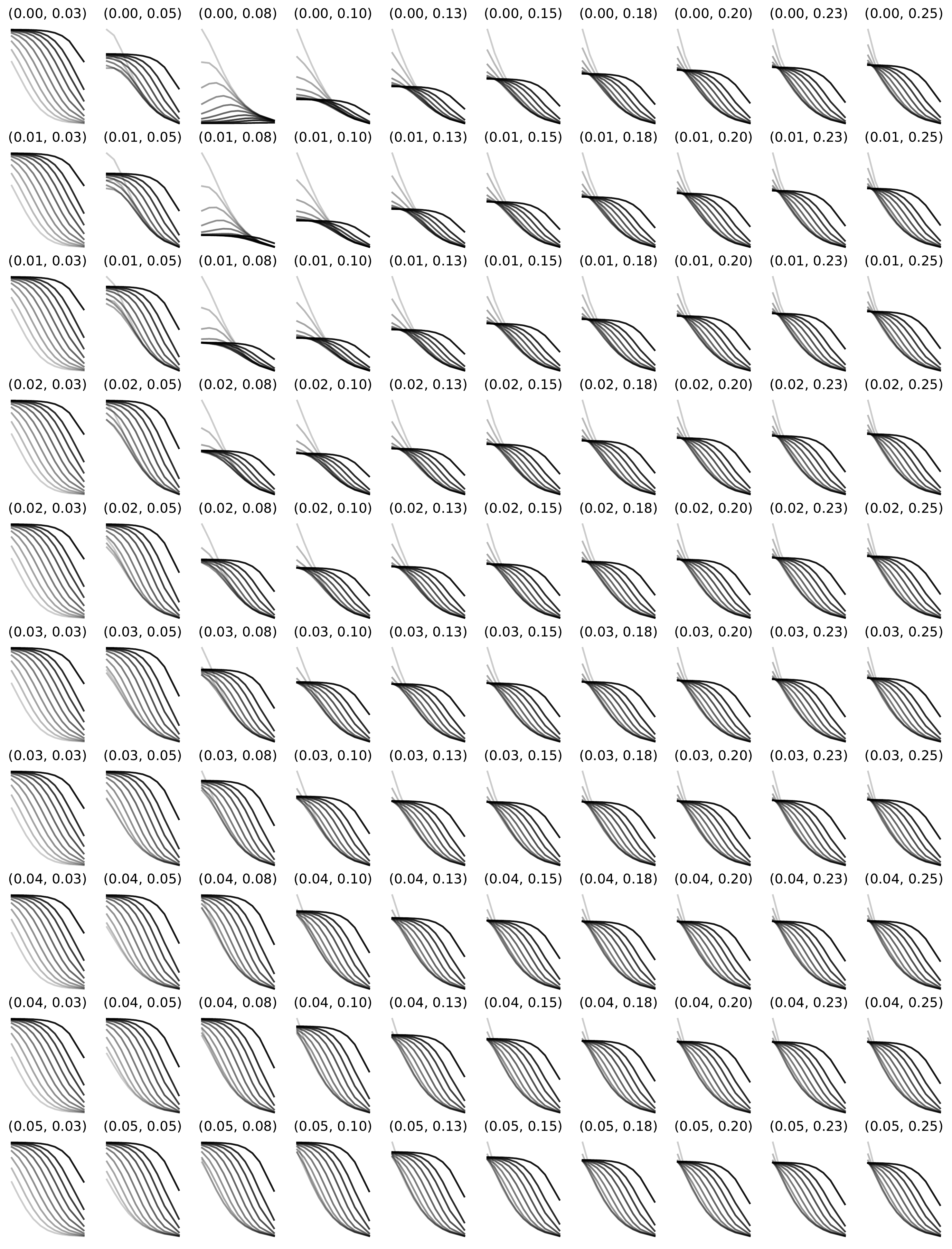}
\caption{Overview of different parametrizations of the model.
    We denote each plot with $(\mu_t - \mu_s, \sigma_t/\sigma_s)$ and report the lower bound $\sqrt{L}$ on the Wasserstein distance.
    Parametrizations in columns four to seven produce qualitatively similar results we observed in our experiments, assuming a linear relationship between the Wasserstein distance and the error rate.
}
\label{fig:simulation}
\end{figure}

Based on choices of the model parameters, the model qualitatively matches our experimental results. We plot different choices in Fig.~\ref{fig:simulation}.

\subsection{Extension to multivariate distributions.}
\label{app:proof-multi}

We now derive a multivariate variant that can be fit to data from a DNN.
Due to the estimation of running statistics in the network, we have access to a diagonal approximation of the true covariance matrix.

We denote the diagonal covariance matrices with matrix elements $\sigma^2_i$ as
\begin{align}
    (\Sigma_t)_{ii} &= (\sigma^2_t)_{i}, \; (\hat{\Sigma}_t)_{ii} = (\hat{\sigma}^2_t)_{i}, \; (\Sigma_s)_{ii} = (\sigma^2_s)_{i} 
\end{align}
and extend our definition of the statistics used for normalization to $\Bar{\mumu}$ and $\Bar{\covd}$:
\begin{align}
    \Bar{\mumu} =& \; \frac{N}{N+n} \mumu_s + \frac{n}{N+n} \hat{\mumu}_t, \; \Bar{\covd} = \frac{N}{N+n} \covd_s + \frac{n}{N+n} \hat{\covd}_t.
\end{align}
The Wasserstein distance between $\Bar{\mumu},\Bar{\covd}$ and $\mumu_t, \covd_t$ is then defined as
\begin{equation}
\begin{aligned}
    W_2^2 =& \; \Tr{\covd_t + \Bar{\covd} - 2 \covd_t^{1/2} \Bar{\covd}^{1/2}} + (\mumu_t - \Bar{\mumu})^T (\mumu_t - \Bar{\mumu}) \\
    &= \sum_{i=1}^D (\sigma^2_t)_i + (\Bar{\sigma}^2)_i - 2 (\Bar{\sigma})_i (\sigma_t)_i + \left((\mu_t)_i - (\Bar{\mu}_t)_i\right)^2  = \sum_{i=1}^D (W_2^2)_i
\end{aligned}
\end{equation}
Every component $(W_2^2)_i$ in the sum above is bounded by the univariate bound discussed above. 
The multivariate Wasserstein distance which sums over the diagonal covariance matrix entries is then bounded by the sums over the individual bounds $L_i$ and $U_i$ given in \eqref{univariatebound}.
\begin{align}
    L_i \leq (W_2^2)_i \leq U_i \Rightarrow \sum_{i=1}^D L_i \leq W_2^2 \leq \sum_{i=1}^D U_i.
\end{align}

\subsection{Limits of Proposition 1}

\paragraph{Limit $n \to \infty$}
In the limit of infinite batch size $n \to \infty$, upper and lower bounds on the expected Wasserstein distance between $\Bar{\mu}, \Bar{\sigma}^2$ and $\mu_t, \sigma^2_t$ both go to zero.
\begin{equation}
\begin{aligned}
    \lim_{n \to \infty} L =& \lim_{n \to \infty}  \left( \sigma_t - \sqrt{ \frac{N}{N+n} \sigma^2_s + \frac{n-1}{N+n} \sigma^2_t } \right)^2 + \frac{N^2}{(N+n)^2} \left(\mu_t -  \mu_s \right)^2 + \frac{n}{(N+n)^2} \sigma_t^2 \\
    =& (\sigma_t - \sigma_t)^2  = 0 \\
    \lim_{n \to \infty} U =&  \lim_{n \to \infty} L + \lim_{n \to \infty} \sigma^5_t \frac{(n-1)}{2(N+n)^2}  \left( \frac{N}{N+n} \sigma^2_s + \frac{1}{N+n} \chi^2_{1-\alpha/2, n-1} \sigma^2_t \right)^{-3/2}  = 0.
\end{aligned}
\end{equation}
The intuition behind this limit is that if a large number of samples from the target domain is given, $\hat{\mu}$ and $\hat{\sigma}^2$ approximate the true target statistics very well. As $\hat{\mu}$ and $\hat{\sigma}^2$ dominate $\Bar{\mu}$ and $\Bar{\sigma}^2$ for large $n$, the expected Wasserstein distance has to vanish.  
\paragraph{Limit $N \to \infty$}
In the opposite limit $N \to \infty$, the expected value of the Wasserstein distance reduces to the Wasserstein distance between source and target statistics.
\begin{align}
    \lim_{N \to \infty} \Bar{\mu} &= \mu_s, \;  \lim_{N \to \infty}  \Bar{\sigma}^2 = \sigma^2_s, \\
    \Rightarrow \;
     \lim_{N \to \infty}  \mathbb{E}[W_2^2] &= \sigma^2_t + \sigma^2_s - 2 \sigma_t \sigma_s + (\mu_t - \mu_s)^2
    = W_2^2 \left( \mu_s, \sigma^2_s, \mu_t, \sigma^2_t \right).
\end{align}

\paragraph{Limiting case $\mu_t = \mu_s$ and $\sigma^2_t = \sigma^2_s$}
When source and target domain coincide, and the statistics $\sigma^2_s = \sigma^2_t$ and $\mu_s = \mu_t$ are known, then the source target mismatch is not an error source.

However, one might assume that source and target domain are different even though they actually coincide. In this case, proceeding with our proposed strategy and using the statistics $\Bar{\mu}$ and $\Bar{\sigma}^2$, the bounds on the expected Wasserstein distance follow from setting $\sigma^2_t$ to $\sigma^2_s$ and $\mu_t$ to $\mu_s$ in Proposition~\ref{prop:bounds}.
\begin{equation}\label{source=target}
\begin{aligned}
    \Bar{\mu} &= \frac{N}{N+n} \mu_t + \frac{n}{N+n} \hat{\mu}_t, \; \Bar{\sigma}^2 = \frac{N}{N+n} \sigma^2_t + \frac{n}{N+n} \hat{\sigma}^2_t, \; L \leq \mathbb{E}[W_2^2] \leq U \\
    L &= \sigma^2_t \left(\frac{2N^2 + 4Nn - N + 2n^2}{(N+n)^2} -2 \sqrt{1 - \frac{1}{N+n}} \right), \\
    U &= L + \sigma^2_t \frac{n-1}{2(N+n)^2} \left(\frac{N + \chi^2_{1-\alpha/2, n-1}}{N+n} \right)^{-3/2}.
\end{aligned}
\end{equation}
It could also be the case that the equality of source and target statistics is known but the concrete values of the statistics are unknown. In our model, this amounts to setting the number of pseudo samples $N$ to zero and assuming that source and target statistics are equal. Setting $N=0$ in equation \eqref{source=target} and keeping $n$ finite yields
\begin{align}
    L &= 2 \sigma^2_t \left(1 - \sqrt{1 - \frac{1}{n}} \right), \; U = L + \sigma^2_t \frac{n-1}{2n^2} \left( \frac{\chi^2_{1-\alpha/2, n-1}}{n} \right)^{-3/2}.
\end{align}

\subsection{Bounds on the normalized Wasserstein distance}

The Wasserstein distance (cf. \textsection\ref{def:wasserstein}) between the interpolating statistics $\Bar{\mu}$, $\Bar{\sigma}^2$ and the target statistics can also be normalized by a factor of $\sigma_s^{-2}$.
Because $\sigma_s^{-2}$ is constant, the bounds on the expectation value of the unnormalized Wasserstein distance discussed in the previous subsections just have to be multiplied by $\sigma_s^{-2}$ to obtain bounds on the normalized Wasserstein distance (cf. \textsection\ref{def:wasserstein-norm}):
\begin{align}
    \frac{L}{\sigma^2_s} \leq \widetilde{W}_2^2 = W_2^2 \left(\frac{\Bar{\mu}}{\sigma_s},, \frac{\Bar{\sigma}^2}{\sigma^2_s},
     \frac{\mu_t}{\sigma_s},
     \frac{\sigma_t^2}{\sigma_s^2} \right) = \frac{1}{\sigma^2_s} W_2^2(\Bar{\mu},
     \Bar{\sigma}^2, \mu_t, \sigma^2_t) \leq \frac{U}{\sigma^2_s}.
\end{align}

\clearpage
\section{Full list of models evaluated on IN}
\label{apx:imagenet_model_descriptions}
The following lists contains all models we evaluated on various datasets with references and links to the corresponding source code.

\subsection{Torchvision models trained on IN} 
Weights were taken from \url{https://github.com/pytorch/vision/tree/master/torchvision/models}

{\small
\begin{enumerate}
\item \texttt{alexnet} \citep{alexnet}
\item \texttt{densenet121} \citep{densenet}
\item \texttt{densenet161} \citep{densenet}
\item \texttt{densenet169} \citep{densenet}
\item \texttt{densenet201} \citep{densenet}
\item \texttt{densenet201} \citep{densenet}
\item \texttt{googlenet} \citep{googlenet} 
\item \texttt{inception\_v3} \citep{inception} 
\item \texttt{mnasnet0\_5} \citep{mnasnet} 
\item \texttt{mnasnet1\_0} \citep{mnasnet} 
\item \texttt{mobilenet\_v2} \citep{mobilenet} 
\item \texttt{resnet18} \citep{he2016resnet} 
\item \texttt{resnet34} \citep{he2016resnet} 
\item \texttt{resnet50} \citep{he2016resnet} 
\item \texttt{resnet101} \citep{he2016resnet} 
\item \texttt{resnet152} \citep{he2016resnet} 
\item \texttt{resnext50\_32x4d} \citep{xie2017aggregated}
\item \texttt{resnext101\_32x8d} \citep{xie2017aggregated} 
\item \texttt{shufflenet\_v2\_x0\_5} \citep{shufflenet} 
\item \texttt{shufflenet\_v2\_x1\_0} \citep{shufflenet} 
\item \texttt{vgg11\_bn} \citep{vgg} 
\item \texttt{vgg13\_bn} \citep{vgg} 
\item \texttt{vgg16\_bn} \citep{vgg} 
\item \texttt{vgg19\_bn} \citep{vgg} 
\item \texttt{wide\_resnet101\_2} \citep{wideresnet} 
\item \texttt{wide\_resnet50\_2} \citep{wideresnet} 
\end{enumerate}
}

\subsection{Robust ResNet50 models}
{\small
\begin{enumerate}
\item \texttt{resnet50 AugMix} \citep{hendrycks2019augmix} \url{https://github.com/google-research/augmix}
\item \texttt{resnet50 SIN+IN} \citep{geirhos2018imagenettrained} \url{https://github.com/rgeirhos/texture-vs-shape}
\item \texttt{resnet50 ANT} \citep{Rusak2020IncreasingTR}
\url{https://github.com/bethgelab/game-of-noise}
\item \texttt{resnet50 ANT+SIN} \citep{Rusak2020IncreasingTR}
\url{https://github.com/bethgelab/game-of-noise}
\item \texttt{resnet50 DeepAugment}
\citep{hendrycks2020many}
\url{https://github.com/hendrycks/imagenet-r}
\item \texttt{resnet50 DeepAugment+AugMix}
\citep{hendrycks2020many}
\url{https://github.com/hendrycks/imagenet-r}

\end{enumerate}
}

\subsection{SimCLRv2 models \citep{chen2020big}}
We used the checkpoints from \url{https://github.com/google-research/simclr} and converted them from TensorFlow to PyTorch with \url{https://github.com/tonylins/simclr-converter}, commit ID: 139d3cb0bd0c64b5ad32aab810e0bd0a0dddaae0.
{\small
\begin{enumerate}
\item \texttt{resnet50} FT100 SK=0 width=1
\item \texttt{resnet101} FT100 SK=0 width=1
\item \texttt{resnet152} FT100 SK=0 width=1
\end{enumerate}
}

\subsection{Robust ResNext models \citep{xie2017aggregated}}
Note that the baseline \texttt{resnext50\_32x4d} model trained on ImageNet is available as part of the \texttt{torchvision} library.
{\small
\begin{enumerate}
\item \texttt{resnext50\_32x4d WSL} \citep{mahajan2018exploring} \url{https://github.com/facebookresearch/WSL-Images/blob/master/hubconf.py}
\item \texttt{resnext101\_32x4d WSL} \citep{mahajan2018exploring} \url{https://github.com/facebookresearch/WSL-Images/blob/master/hubconf.py}
\item \texttt{resnext101\_32x8d Deepaugment+AugMix} \citep{hendrycks2020many} \url{https://github.com/hendrycks/imagenet-r}
\end{enumerate}
}

\subsection{ResNet50 with Group Normalization \citep{wu2018group}} 
Model weights and training code was taken from \url{https://github.com/ppwwyyxx/GroupNorm-reproduce}
{\small
\begin{enumerate}
\item \texttt{resnet50 GroupNorm}
\item \texttt{resnet101 GroupNorm}
\item \texttt{resnet152 GroupNorm}
\end{enumerate}
}

\subsection{ResNet50 with Fixup initialization \citep{zhang2019fixup}} 
Model weights and training code was taken from \url{https://github.com/hongyi-zhang/Fixup/tree/master/imagenet}.
For training, we keep all hyperparameters at their default values and note that in particular the batchsize of 256 is a sensitive parameter.

{\small
\begin{enumerate}
\item \texttt{resnet50 FixUp}
\item \texttt{resnet101 FixUp}
\item \texttt{resnet152 FixUp}
\end{enumerate}
}

\end{document}